\documentclass[journal]{IEEEtai}
\pdfoutput=1
\usepackage[colorlinks,urlcolor=blue,linkcolor=blue,citecolor=blue]{hyperref}

\usepackage{color,array}

\usepackage{graphicx}

\usepackage{algorithm}
\usepackage{algorithmicx}
\usepackage{algpseudocode}
\usepackage{amsmath,amsfonts}
\usepackage{booktabs} 
\usepackage{multirow}
\usepackage{xcolor}         
\usepackage{threeparttable}
\usepackage{caption}
\newcommand{\update}[1]{{\textcolor{black}{#1}}}
\newcommand{\boldres}[1]{{\textbf{\textcolor{red}{#1}}}}
\newcommand{\secondres}[1]{{\underline{\textcolor{blue}{#1}}}}
\usepackage{colortbl} 
\usepackage{bbding}
\usepackage{subfigure}

\begin{document}

\title{DC-Mamber: A Dual Channel Prediction Model based on Mamba and Linear Transformer for Multivariate Time Series Forecasting}

\author{Bing Fan$^{\dag}$, Shusen Ma$^{\dag}$, Yun-Bo Zhao$^{\ast}$, \IEEEmembership{Senior Member, IEEE}, and Yu Kang, \IEEEmembership{Senior Member, IEEE}
\thanks{This work was supported by the National Natural Science Foundation of China (No. 62173317). $^{\dag}$These authors contributed equally to this work.}
\thanks{Bing Fan$^{\dag}$ is with the  School of Computer Science and Technology, University of Science and Technology of China, Hefei, Anhui, China.}
\thanks{Shusen Ma$^{\dag}$ is with the Institute of Advanced Technology, University of Science and Technology of China, Hefei, Anhui, China.}
\thanks{Yun-Bo Zhao$^{\ast}$ (corresponding author) is with the Department of Automation, University of Science and Technology of China, Hefei, Anhui, China, the Institute of Advanced Technology, University of Science and Technology of China, Hefei, Anhui, China, and also with the Institute of Artificial Intelligence, Hefei Comprehensive National Science Center, Anhui, China (e-mail: ybzhao@ustc.edu.cn).}
\thanks{Yu Kang is with the Institute of Advanced Technology, University of Science and Technology of China, Hefei, Anhui, China, the Department of Automation, University of Science and Technology of China, Hefei, Anhui, China, and also with the Institute of Artificial Intelligence, Hefei Comprehensive National Science Center, Anhui, China.}
}


\maketitle

\begin{abstract}
In multivariate time series forecasting (MTSF), existing strategies for processing sequences are typically categorized as channel-independent and channel-mixing. The former treats all temporal information of each variable as a token, focusing on capturing local temporal features of individual variables, while the latter constructs a token from the multivariate information at each time step, emphasizing the modeling of global temporal dependencies. 
Current mainstream models are mostly based on Transformer and the emerging Mamba. 
Transformers excel at modeling global dependencies through self-attention mechanisms but exhibit limited sensitivity to local temporal patterns and suffer from quadratic computational complexity, restricting their efficiency in long-sequence processing. In contrast, Mamba, based on state space models (SSMs), achieves linear complexity and efficient long-range modeling but struggles to aggregate global contextual information in parallel.
To overcome the limitations of both models, we propose DC-Mamber, a dual-channel forecasting model based on Mamba and linear Transformer for time series forecasting. Specifically, the Mamba-based channel employs a channel-independent strategy to extract intra-variable features, while the Transformer-based channel adopts a channel-mixing strategy to model cross-timestep global dependencies. DC-Mamber first maps the raw input into two distinct feature representations via separate embedding layers. These representations are then processed by a variable encoder (built on Mamba) and a temporal encoder (built on linear Transformer), respectively. Finally, a fusion layer integrates the dual-channel features for prediction. Extensive experiments on eight public datasets confirm DC-Mamber’s superior accuracy over existing models.
\end{abstract}


\section{Introduction}

\IEEEPARstart{A}time series is a sequence of data points ordered chronologically, containing local and global information. Effectively extracting and learning these temporal features is crucial for understanding the developmental patterns of various processes. In particular, for MTSF, capturing these temporal patterns can significantly enhance prediction accuracy. MTSF has been widely applied in the real world, such as in traffic flow forecasting \cite{9226432, karim2024probabilistic, qi2025routeformer}, weather forecasting \cite{zenkner2023flexible}, power forecasting \cite{almaghrabi2022solar} and so on. In these applications, forecasting the future states of target variables enables proactive strategy formulation, thereby maximizing benefits or minimizing losses \cite{ma2024multivariate}.

In MTSF, current data processing strategies are typically categorized into channel-mixing \cite{zhou2021informer} and channel-independent \cite{ma2024multivariate, liu2024itransformer, Yuqietal-2023-PatchTST}. The channel-mixing strategy treats multivariate data at each timestamp as a token, while the channel-independent strategy regards each variable’s full temporal information as a token. Currently, Transformer-based models \cite{zhou2021informer, zhang2022crossformer, ma2024multivariate, liu2024itransformer} dominate MTSF due to their capacity for modeling global dependencies through attention mechanisms. However, regardless of the strategy employed, canonical Transformer-based models face computational bottlenecks: when using channel-mixing, the computational complexity grows quadratically with sequence length, whereas with channel-independence, it grows quadratically with the number of variables \cite{ma2024multivariate}.
Additionally, although Transformer-based models are well-suited for capturing global dependencies, they often underperform in modeling local dependencies. This shortcoming reduces their sensitivity to local patterns such as short-term trends and periodicities, ultimately leading to deviations between predicted and actual values.

Mamba \cite{Gu2023MambaLS6}, as a strong competitor to Transformer \cite{NIPS2017_3f5ee243}, leverages SSMs \cite{Gu2021S4} to process time series data. By introducing an innovative parallel scan mechanism, it transforms traditional recursive operations into efficient convolutional computations. Unlike the global interaction mechanism employed by Transformers \cite{zhang2022crossformer, liu2024itransformer}, Mamba exhibits notable advantages in capturing local information, making it particularly effective at capturing local features within sequences—an essential capability for MTSF tasks. However, the information flow in Mamba primarily relies on the hidden states of the SSMs, where each current hidden state is determined jointly by the previous hidden state and the current input \cite{Gu2021S4}. 
This mechanism prevents Mamba from explicitly modeling direct connections between different tokens like Transformer does, thereby presenting certain limitations in capturing global dependencies within the input.

Through the above analysis, we can observe that Mamba and Transformer-based models may exhibit complementary characteristics when applied to MTSF tasks \cite{wang2025S-Dmamba,ahamed2024timemachine,ZHANG2025129429}. Specifically, Mamba can compensate for the Transformer's shortcomings in local dependencies modeling, while Transformers can enhance the extraction of global dependencies, which Mamba lacks. Furthermore, experimental observations indicate that both channel-mixing and channel-independent strategies positively impact model performance to varying degrees. We attribute this to their respective strengths: the channel-mixing strategy facilitates the modeling of global temporal features \cite{zhou2021informer}, whereas the channel-independent strategy excels at capturing intra-variable local patterns \cite{Yuqietal-2023-PatchTST}. It is evident that the characteristics of these two strategies align well with the respective properties demonstrated by the Transformer and Mamba in MTSF. This suggests that the channel-mixing strategy may be more suitable for Transformer, while the channel-independent approach could be a better fit for Mamba architectures.
To maintain computational efficiency, we propose employing a linear Transformer as the core framework for the channel-mixing component, ensuring that the overall model achieves linear-time complexity.

Therefore, to simultaneously consider both local and global information and reduce the computational complexity of the model, we have proposed a \textbf{d}ual-\textbf{c}hannel prediction model based on \textbf{Mamb}a and Linear Transform\textbf{er}, named DC-Mamber. Technically, DC-Mamber first passes the original input sequences through the V-Embedding and T-Embedding layers for feature space mapping. Then the outputs of the V-Embedding and T-Embedding layers are  fed into the V-Encoder and T-Encoder, respectively. Here, the V-Encoder refers to an encoder layer based on the channel-independent strategy and Mamba, while the T-Encoder is built upon the channel-mixing strategy and Transformer. Finally, the outputs from the V-Encoder and T-Encoder are fused through a Feature-fusion layer, and the fused representation is projected through a Projection layer to generate the final predictions.
The main contributions of this paper are summarized as follows:
\begin{itemize}
    \item We propose a novel dual-channel prediction model, DC-Mamber, based on Mamba and linear attention mechanism. DC-Mamber is capable of simultaneously capturing local temporal dependencies within variables and global temporal dependencies across the entire input, significantly enhancing the model’s representational capacity. 
    \item We design dedicated channel architectures for both channel-independent and channel-mixing strategies, carefully considering their characteristic differences. Experimental results demonstrate that both designs effectively enhance model performance.
    \item We conduct extensive experiments on eight publicly available datasets. The results show that our proposed model outperforms state-of-the-art(SOTA)  baselines, achieving an average reduction of 4.2\% in MSE and 4.9\% in MAE metrics. 
\end{itemize}

\section{Related Work}

\subsection{Transformer-based Methods}
Due to the outstanding performance of Transformers in the fields of natural language processing (NLP) \cite{devlin2019bert} and computer vision (CV) \cite{zhudeformable}, many studies have explored how to adapt Transformer-based models to MTSF tasks. Informer \cite{zhou2021informer} proposes a generative style decoder that can achieve single-shot multi-step forecasting, effectively avoiding the accumulation of errors associated with step-by-step forecasting. Stationary Transformer \cite{NEURIPS2022_4054556f} addresses the over-stationarization issue by proposing De-stationary Attention. By utilizing seasonal-trend decomposition, FEDformer \cite{zhou2022fedformer} significantly minimizes the distribution discrepancy between input and output sequences while incorporating a frequency-domain attention mechanism to to bolster its resilience against noise. TimesNet \cite{wu2022timesnet} captures complex periodic patterns by reshaping one-dimensional time series into two-dimensional tensors and leverages TimesBlocks to adaptively discover multiple periodicities and extract temporal features.

However, all of the above methods adopt channel-mixing strategies, which may lead to information coupling issues \cite{ma2024multivariate}. To address this limitation, several recent studies have proposed channel-independent strategies \cite{ma2024multivariate, Yuqietal-2023-PatchTST, liu2024itransformer}. PatchTST proposes feeding each variable independently into a shared-parameter backbone. PCDformer  \cite{ma2024multivariate} introduces parallel convolutional layers to process different variables separately, reducing intra-variable  interference and thereby enhancing model robustness. iTransformer \cite{liu2024itransformer} reconfigures Transformer-based frameworks for MTSF by employing an inverted structural design to emphasize the autonomy of variables.
 Although these methods treat each variable as an independent token \cite{ma2024multivariate, liu2024itransformer}, their computational complexity scales quadratically with the total number of input variables, which limits their scalability to scenarios involving a large number of variables.

\subsection{Mamba-based Methods}
Due to its efficient sequence modeling capabilities and its ability to capture temporal dependencies, Mamba has demonstrated strong performance in handling time series data. Timemachine \cite{ahamed2024timemachine} is the first model to apply Mamba to MTSF, while simultaneously adopting both channel-independent and channel-mixing strategies. Bi-Mamba+ \cite{liang2024bi} introduces a Series-Relation-Aware decider that automatically selects between the channel-independent and channel-mixing strategies. S-Mamba \cite{wang2025S-Dmamba} proposes a bidirectional Mamba architecture to mitigate Mamba’s unilateral nature, which cannot attend to global variables like Transformer. MixMamba \cite{alkilane2024mixmamba} leverages Mamba as an expert within a mixture-of-experts (MoE) system to fully exploit its strengths in content-based reasoning, enabling different experts to learn different features and thereby enhancing model robustness.

However, these methods mainly focus on exploiting Mamba’s selective mechanism to capture temporal dependencies, while lacking exploration of the potential synergy between Mamba and Transformer architectures. Although FMamba \cite{ma2024fmamba} combines Mamba with a linear Transformer, it only adopts the channel-independent strategy, without exploring the benefits of a hybrid strategy.

\subsection{Other Methods}
In addition to Transformer-based and Mamba-based forecasting models, other common approaches include models based on RNN \cite{salazar2025distance} and their variants such as LSTM \cite{ma2023tcln} and GRU \cite{jhin2024addressing}, as well as models based on CNN and MLP. RNN leverage their recurrent structure to memorize historical information, utilizing both the current input and the hidden states to predict future outputs. However, during the training of RNN, as the time steps unfold, gradients are propagated through continuous multiplications in backpropagation. This mechanism can cause gradients to either exponentially grow (gradient explosion) or exponentially vanish (gradient vanishing) due to repeated multiplications with the weight matrices. As a result, compared to Transformer or Mamba, RNN are less capable of handling long sequences.

Unlike the recurrent structure of RNN, CNN extract local features from time series data through convolutional kernels and leverage their translation invariance to capture temporal dependencies.For instance, LSTNet \cite{lai2018modeling} initially extracts short-term local dependencies between variables using convolutional layers, and then captures long-term patterns in the input series through RNN layers. PCDformer \cite{ma2024multivariate} uses the Inception structure \cite{Szegedy2015} to capture the multi-scale temporal relations, which can improve the model's representation ability.
However, due to the limited receptive field of convolutional kernels, CNN struggle to effectively capture long-range dependencies.Recent studies have shown that simple MLP-based forecasting models can achieve or even surpass the performance of previous Transformer-based models. For instance, DLinear \cite{3-DLinear} demonstrates through experiments that MLP-based models not only achieve SOTA results but also avoid complex computations compared to Transformer-based models. RLinear \cite{li2023revisiting} verifies that the channel-independent strategy has a significant impact on overall prediction performance.
Nevertheless, unlike RNN, which can directly model temporal dependencies through recurrent mechanisms, and CNN, which can capture temporal patterns via convolutional kernels, MLPs rely solely on fully connected layers for input processing, and thus lacking relatively explicit temporal modeling capabilities.

\section{Methodology}

\subsection{Problem Statement}
In MTSF task, given a dataset
$\mathbf{X}_{data} \in \mathbb{R}^{T \times V}$, where $T$ denotes the total length of the time series and $V$ denotes the number of variables within the sequence.
We aim to predict future $W$ time steps
$\mathbf{X}=\left\{\mathbf{x}_{L+1}, \ldots, \mathbf{x}_{L+W}\right\} \in \mathbb{R}^{W \times V}$ 
based on the historical sequence $\mathbf{X}=\left\{\mathbf{x}_{1}, \ldots, \mathbf{x}_{L}\right\} \in \mathbb{R}^{L \times V}$ with a fixed-length $L$.
To effectively capture both the global temporal dependencies and the intra-variable temporal dependencies, we propose a dual-channel representation method, which adopts both channel-independent and channel-mixing strategies to process the input data. Specifically, the input fed into the model can be represented in two forms: time series tokens and variable tokens, denoted as
$\mathbf{X}_{t-token} \in \mathbb{R}^{L \times V}$ and
$\mathbf{X}_{v-token} \in \mathbb{R}^{V \times L}$, respectively.
For channel-mixing, the input is organized in the form of time series tokens, where the token at the $i$-th time step is defined as
$\boldsymbol{x}_{1: V}^{(i)}=\left(x_{1}^{(i)}, \ldots, x_{V}^{(i)}\right)$, with $i = 1, \ldots, L$.
For channel-independence, the input is represented as variable tokens, where the token for the $j$-th variable is defined as
$\boldsymbol{x}_{1: L}^{(j)}=\left(x_{1}^{(j)}, \ldots, x_{L}^{(j)}\right)$, with $j = 1, \ldots, V$.

\subsection{The Structure of DC-Mamber}
\begin{figure*}[htbp]
  \centering
  \includegraphics[width=1.03\textwidth]{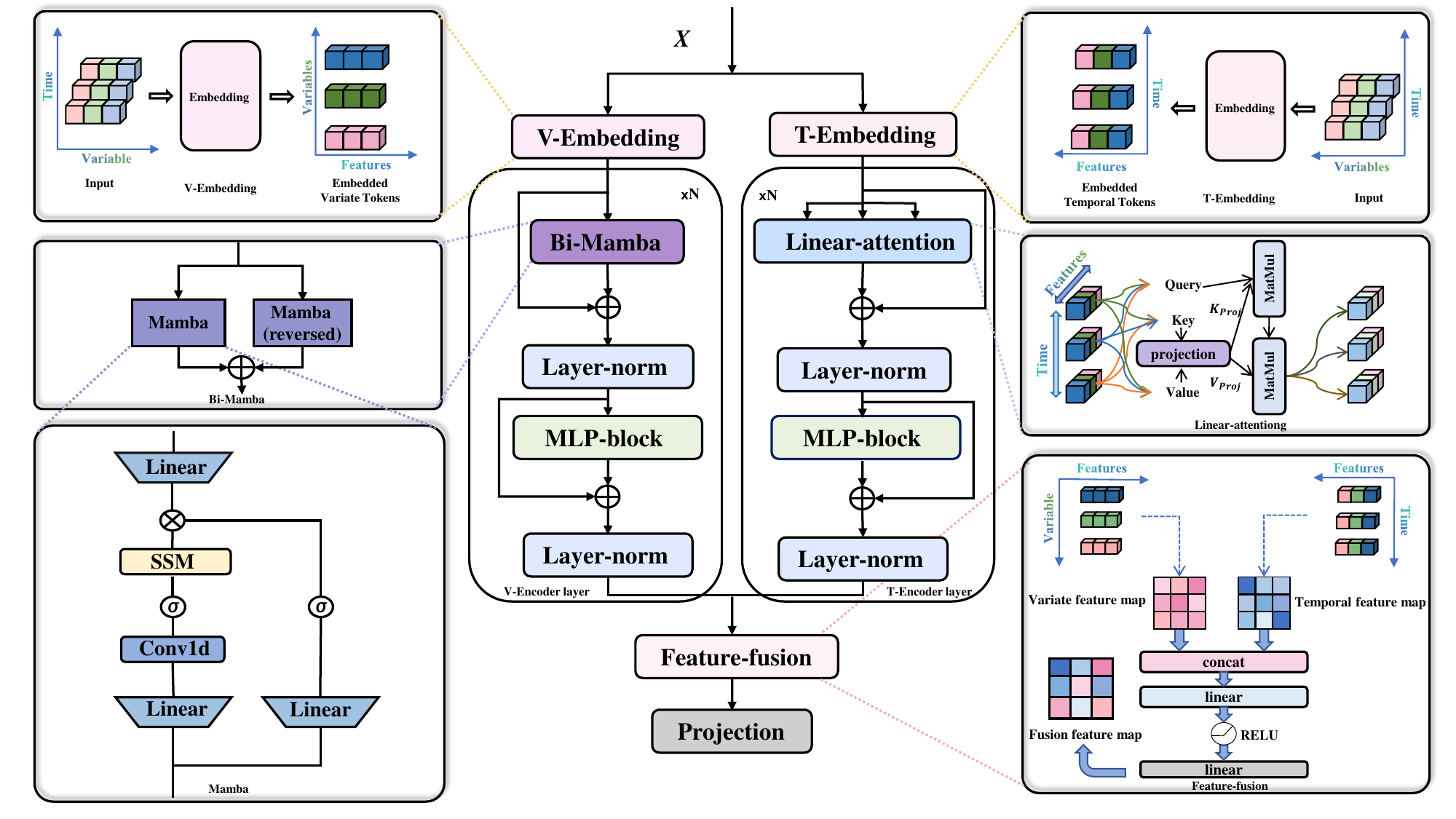} 
  \captionsetup{width=\linewidth}
  \caption{Model structure}
  \label{fig:1-Model_sturcture}
\end{figure*}

The overall framework of DC-Mamber adopts a dual-stream architecture design, with its core components comprising: 1) a Dual-Channel Token Embedding Layer for processing the input, including temporal token embedding (T-Embedding) and variable token embedding (V-Embedding); 2) an $N$-layer stacked dual-channel encoder module, consisting of a temporal encoder (T-Encoder) to extract global temporal dependencies and a variable encoder (V-Encoder) to capture local temporal dependencies within individual variables; 3) a Feature-fusion module for processing temporal and variable features extracted in a decoupled manner. The detailed model architecture is shown in Figure \ref{fig:1-Model_sturcture}. DC-Mamber is capable of capturing both global temporal and local intra-variable  multi-level features simultaneously. Moreover, this decoupled feature extraction strategy prevents cross-feature interference, ensuring that the model can extract features from temporal and variable dimensions independently without interference from the other pattern.

\subsubsection{Dual-Channel Token Embedding Layer}  
In MTSF task, data representation directly influences the model's ability to capture dependencies across temporal and variable dimensions. Existing methods typically adopt two strategies: one is the channel-mixing strategy, which treats values of all variables at each timestamp as a single token \cite{zhou2021informer}; the other is the channel-independent strategy, which considers values of each variable across all timestamps as a single token \cite{Yuqietal-2023-PatchTST}. However, both strategies exhibit inherent limitations in modeling dependencies in temporal and variable dimensions. Specifically, the channel-mixing strategy embeds all variables with different physical meanings at the same timestamp into a single token, leading to intra-variable information entanglement and hindering the model's ability to capture the temporal dynamics within individual variables \cite{ma2024multivariate, liu2024itransformer}. On the other hand, the channel-independent strategy struggles to capture relationships between variables at the same time point, often overlooking instantaneous intra-variable correlations. To address these limitations, we propose a Dual-Channel Token Embedding Layer that combines the advantages of both strategies. Unlike the coupled feature extraction process in TSMixer \cite{40-TSMixer}, we decouple the feature extraction process of temporal tokens and variable tokens, enabling simultaneous modeling of intra-variable temporal dynamics and intra-variable relationships. As illustrated in Figure~\ref{fig:1-Model_sturcture}, the core idea of the Dual-Channel Token Embedding Layer is to generate dual-stream high-dimensional embeddings through parallel linear transformations, including T-Embedding and V-Embedding. This design not only explicitly models dependencies in both temporal and variable dimensions but also provides rich, high-dimensional representations for the subsequent T-Encoder and V-Encoder. The implementation details are as follows:

\textbf{T-Embedding}
To implement the channel-mixing strategy, we first introduce positional encoding (PE) \cite{NIPS2017_3f5ee243} to the original input data $\mathbf{X}\in \mathbb{R}^{B\times L \times V}$. The purpose of PE is to embed temporal information into the data, thereby preserving the sequential continuity. After applying PE, we project the variable dimension $V$ to a high-dimensional feature space $D$, generating the temporal embedding token: $\mathbf{X}_{tem} \in \mathbb{R}^{B\times L \times D}$, where $B$ denotes the batch size dimension. This process preserves the instantaneous correlations among variables within the time series and provides the necessary feature representation for the T-Encoder to extract global temporal dependencies. The detailed computation process is as follows: 
\begin{equation}
 \text{T-Embedding}(x) = \text{ReLU}(\mathbf{W}_{1}{x} + \mathbf{b}_1)\mathbf{W}_2 + \mathbf{b}_2 ,
\end{equation}where \(\mathbf{W}_1 \) and \(\mathbf{W}_2 \) are learnable weight matrices, \(\mathbf{b}_1\) and \(\mathbf{b}_2\) are bias terms, \(\text{ReLU(.) }\) denotes the rectified linear unit activation function introducing nonlinear mapping, and $x$ represents the input data after PE processing.

\textbf{V-Embedding}
To implement the channel-independent strategy, we first  permute the dimensions of the original input data $\mathbf{X}\in \mathbb{R}^{B\times L \times V}$ to obtain $\mathbf{X^T}\in \mathbb{R}^{B\times V \times L}$. This transposition operation effectively treats each variable as an independent channel. Subsequently, we employ a linear layer to map the temporal dimension $L$ to a high-dimensional feature space $D$, thereby generating the variable embedding token: $\mathbf{X}_{var} \in \mathbb{R}^{B\times V \times D}$. This process preserves the temporal dependencies within individual variables, enabling the explicit extraction of their dynamic temporal evolution patterns during feature processing. It also prepares the representation for the V-Encoder to model both local temporal dependencies and intra-variable  relationships. The detailed computational procedure is formulated as:
\begin{equation} 
\begin{split} 
\text{V-Embedding}(x) &= \text{ReLU}(\mathbf{W}{1}\cdot\text{permute}(x))\mathbf{W}_2 \\&+\text{ReLU}(\mathbf{b}_1)\mathbf{W}_2+ \mathbf{b}_2 , 
\end{split} 
\end{equation} 
where \(\mathbf{W}_1 \) and \(\mathbf{W}_2 \) are learnable weight matrices,  \(\mathbf{b}_1\) and \(\mathbf{b}_2\) are bias terms, \(\text{ReLU(.) }\)is the activation function that introduces nonlinearity, permute$(.)$denotes the transposition operation,
and $x$ refers to the original input data.

\subsubsection{T-Encoder}
In MTSF task, capturing global temporal dependencies is crucial for accurate forecasting. We employ  T-Encoder to learn such dependencies. The linear attention mechanism in the T-Encoder receives the temporal embedding tokens $\mathbf{X}_{tem}$ processed by the T-Embedding, enabling simultaneous attention to information across all temporal tokens. This module consists of $N$ stacked T-Encoder layers, each of which integrates a linear attention mechanism, hierarchical residual connections, MLP-blocks, and layer normalization to efficiently extract dependencies along the temporal dimension. The specific process is as follows:

First, the input data is processed by an attention mechanism to extract temporal dependencies. However, traditional self-attention mechanisms suffer from a computational complexity of $O(L^2)$, which is inefficient for long-sequence time series data. Inspired by Linformer \cite{wang2020linformer}, we adopt a linear attention mechanism that reduces the complexity to $O(Lk)$ by using projection matrices $E,F\in\mathbb{R}^{k\times L}$ to project the Key and Value matrices into lower-dimensional representations , where $k \ll L$. Specifically, given $Q,K,V\in\mathbb{R}^{L\times d}$(where $d$ is the embedding dimension), the projected Key and Value matrices become:$\quad K_{\mathrm{proj}}=EK\in\mathbb{R}^{k\times d},\quad V_{\mathrm{proj}}=FV\in\mathbb{R}^{k\times d}$.
Therefore, the final attention computation is simplified as:
\begin{equation}
\begin{split} 
&\text{Linear-attention}(Q,K_{\mathrm{proj}},V_{\mathrm{proj}}) \\&=\mathrm{softmax}\left(\frac{QK_{\mathrm{proj}}^{T}}{\sqrt{d}}\right)V_{\mathrm{proj}}.
\end{split} 
\end{equation}

In this way, the linear attention mechanism retains the expressiveness of traditional attention while significantly reducing computational cost, making it suitable for long-sequence time series modeling.

Next, hierarchical residual connections are employed to alleviate the vanishing gradient problem in deep networks, encourage feature reuse, and ensure that low-level information is preserved in deeper layers. In addition, to reduce the model’s dependency on the distribution of input data, layer normalization is introduced to enhance generalization and improve the model’s representational capacity. Specifically, the computation in each T-Encoder layer is as follows:
\begin{equation}
\begin{split}
   & X_{0} = {X}_{tem} +\mathrm{Dropout}(\text{Linear-attention}({X}_{tem})), \\  
   & X_{1} = \text{Layer-norm}(X_{0}), \\
   & Y_{T-Encoder} = \text{Layer-norm}\left(X_{1}+\text{MLP-block}(X_{1})\right) ,  
\end{split}
\end{equation}
where the MLP-block is implemented using 1D convolutional layers, as detailed below:



\begin{equation}
\begin{split}
Z_1 &= \bigl[\mathrm{Dropout}\bigl(\mathrm{Conv1D}\bigl(\mathrm{RELU}\bigl( \\
&\quad \mathrm{Dropout}\bigl(\mathrm{Conv1D}\bigl(X_{1}^{T}\bigr)\bigr)\bigr)\bigr)\bigr)\bigr]^T.
\end{split}
\end{equation}

Through the channel-mixing strategy, T-Encoder is capable of capturing global temporal dependency features in time series sequences. These features are subsequently combined with the output of the V-Encoder and fed into the Feature-fusion module to further model complex dependencies within the time series.

\subsubsection{V-Encoder}
In addition to the global temporal dependencies captured by the T-Encoder, it is also crucial for the model to consider the temporal evolution patterns within each variable for MTSF. The V-Encoder receives the variable embedding tokens $\mathbf{X}_{var}$ produced by the V-Embedding and leverages bidirectional Mamba (Bi-Mamba) to capture local temporal dependencies within each variable. Unlike the T-Encoder which focuses on global modeling along the temporal dimension, the V-Encoder's core feature extraction module is based on Bi-Mamba. It maintains architectural symmetry with the T-Encoder to form a complementary structure for temporal modeling. The V-Encoder layer is computed as follows: 
\begin{align}
\begin{split}
& X_{0}' ={X}_{var}+\mathrm{Dropout}(\text{Bi-Mamba}({X}_{var})),\\
& X_{1}'=\text{Layer-norm}(X_{0}') , \\
&Y_{V-Encoder}=\text{Layer-norm}\left(X_{1}'+\text{MLP-block}(X_{1}')\right) .
\end{split}
\end{align}
Due to its sequential processing nature, the conventional Mamba struggles to fully exploit contextual information, limiting its ability to model intra-variable  relationships under the channel-independent strategy. Inspired by S-Mamba \cite{wang2025S-Dmamba}, the V-Encoder adopts Bi-Mamba. This structure uses a bidirectional scanning mechanism to ensure that the model has access to the complete information from other variables when processing each individual variable. The following equations illustrate the computation in this bidirectional structure: 
\begin{align}
\begin{split}
& \overrightarrow{\boldsymbol{Y}}=\overrightarrow{\text{Mamba}}({X_{var}}),\\
& \overleftarrow{\boldsymbol{Y}}=\overleftarrow{\text{Mamba}}({X_{var}}),\\
& Y_\text{Bi-Mamba}=\overrightarrow{Y}+\overleftarrow{Y} ,
\end{split}
\end{align}
Mamba represents all recurrent processes with hidden states through two sets of equations, as described in Algorithm \ref{The process of Mamba Block}. Specifically, the SSMs maps an input function $x(t)$ to an output function $y(t)$ through a set of first-order differential equations describing the evolution of the system's latent state $h(t)$, and another equation that links the relationship between hidden state $h(t)$ and output $y(t)$. The mathematical form is given by: 
\begin{align}
\begin{split}
h(t)^{\prime} & =\boldsymbol{A} h(t)+\boldsymbol{B} x(t),\\
y(t) & =\boldsymbol{C} h(t) ,
\end{split}
\end{align}
where $\boldsymbol{A} \in \mathbb{R}^{N \times N}, \boldsymbol{B} \in \mathbb{R}^{N \times D},\boldsymbol{C} \in \mathbb{R}^{N \times D}$ are learnable parameter matrices. The continuous parameters $\boldsymbol{A}$ and $\boldsymbol{B}$ can be discretized to enable the model to handle real-world tasks. Among various discretization methods, the zero-order hold method is one common approach, which requires a fixed time interval $\Delta$ to perform discretization. The discrete parameters are given by:
\begin{align}
\begin{split}
\overline{\boldsymbol{A}}&=\exp (\Delta \boldsymbol{A}), \\
\overline{\boldsymbol{B}}&=(\Delta \boldsymbol{A})^{-1}(\exp (\Delta \boldsymbol{A})-\boldsymbol{I}) \cdot \Delta \boldsymbol{B} .
\end{split}
\end{align}
The resulting discrete SSMs equations are: 
\begin{align}
\begin{split}
h_{t} &= \overline{\boldsymbol{A}} h_{t-1} + \overline{\boldsymbol{B}} x_{t}, \\
y_{t} &= \boldsymbol{C} h_{t}.  
\end{split}
\end{align}

\renewcommand{\algorithmicrequire}{\textbf{Input:}}
\renewcommand{\algorithmicensure}{\textbf{Output:}}

\begin{algorithm}[H]

\caption{The process of Mamba Block}
\label{The process of Mamba Block}
\begin{algorithmic}[1]
\Require $\mathbf{X} : (B, V, D)$
\Ensure $\mathbf{Y} : (B, V, D)$
\State $\mathbf{x}, \mathbf{z} : (B, V, ED) \gets \text{Linear}(X)$\{Linear projection\} 
\State $\mathbf{x}' : (B, V, ED) \gets \text{SiLU}(\text{Conv1D}(\mathbf{x}))$
\State $\mathbf{A} : (D, N) \gets \text{Parameter}$ \{Structured state matrix\}
\State $\mathbf{B}, \mathbf{C} : (B, V, N) \gets \text{Linear}(\mathbf{x}'), \text{Linear}(\mathbf{x}')$
\State $\Delta : (B, V, D) \gets \text{Softplus}(\text{Parameter} + \text{Broadcast}(\text{Linear}(\mathbf{x}')))$
\State ${\overline{A},\overline{B}} : (B, V, D, N) \gets \text{discretize}(\Delta, \mathbf{A}, \mathbf{B})$ \{Input-dependent parameters and discretization\}
\State $\mathbf{y} : (B, V, ED) \gets \text{SelectiveSSM}(\overline{A},\overline{B}, \mathbf{C})(\mathbf{x}')$
\State $\mathbf{y}' : (B, V, ED) \gets \mathbf{y} \otimes \text{SiLU}(\mathbf{z})$
\State $\mathbf{Y} : (B, V, D) \gets \text{Linear}(\mathbf{y}')$ \{Linear Projection\}
\end{algorithmic}

\end{algorithm}

Discretized SSMs support two computational modes: linear recurrence for inference and global convolution for training. The linear recurrence mode offers efficient inference but lacks parallelism, while the convolutional mode enables parallel computation but cannot handle infinitely long contexts. The LSSL model \cite{Gu2021LSSLs} flexibly integrates recurrence, convolution, and continuous-time modeling, enabling conversions among the three modes.

S4 \cite{Gu2020HiPPO} introduced the HiPPO matrix to address long-range dependencies. Building on S4, the Mamba model further integrates the selective scan algorithm and hardware-aware algorithm \cite{Gu2023MambaLS6}, enabling dynamic filtering of irrelevant information and improving computational efficiency through parallel scan, kernel fusion, and recomputation techniques. These enhancements significantly accelerate intermediate computations while preserving model expressiveness.

\subsubsection{Feature-fusion}
The decoupled temporal features extracted by the dual-channel encoders play a critical role in forecasting tasks. To help the model better utilize the heterogeneous feature information derived from the channel-independent and channel-mixing strategies, we propose a linear Feature-fusion architecture. This module integrates the features extracted by the dual-channel encoders to comprehensively capture both global temporal dependencies and local intra-variable temporal dependencies.

Specifically, after extracting features along the temporal and variable dimensions of the time series respectively, the T-Encoder and V-Encoder generate a temporal feature map ${M_{tem}} \in \mathbb{R}^{B \times L \times D}$ and a variable feature map ${M_{var}} \in \mathbb{R}^{B \times V \times D}$, where $B$ denotes batch size, $L$ is sequence length, $V$ is number of variables, and $D$ is feature dimension. Due to the dimensional mismatch between the two feature maps, we align the temporal features with the variable dimension before fusion. This is achieved by applying a linear projection of $M_{tem}$ along temporal dimension $L$  to match variable dimension $V$, keeping the feature dimension $D$ unchanged. This yields an aligned temporal feature map, denoted as $M_{tem}' \in \mathbb{R}^{B \times V \times D}$.

After alignment, we concatenate the transformed temporal feature map $M_{tem}'$ and the variable feature map $M_{var}$ along the feature dimension to form a joint feature map that integrates both types of information. Finally, the joint feature map undergoes further integration through MLP. The specific computation of the Feature-fusion is given by:
\begin{equation}
\begin{split}
&\text{Feature-fusion}(M'_{\text{tem}},M_{\text{var}}) \\&
= \text{Layer-Norm}\Bigl(
    \mathbf{W}_2 \bigl( \text{ReLU}\bigl(
    \mathbf{W}_1 x + \mathbf{b}_1
    \bigr)\bigr) + \mathbf{b}_2
\Bigr).
\end{split}
\end{equation}
where, $\mathbf{W}_1\in\mathbb{R}^{2D\times D},\mathbf{W}_2\in\mathbb{R}^{D\times D}$ are learnable parameter matrices, ReLU$(.)$ serves as the activation function to introduce non-linear representation capability to the model, and the input $x\in\mathbb{R}^{B\times V\times 2D}$ represents the concatenated feature map of $M_{tem}'$ and $M_{var}$.

\section{Experimental Results}

\subsection{Datasets and Baselines}
1) \textbf{Datasets.} In this study, we selected multiple representative real-world datasets from diverse domains to comprehensively evaluate model performance. Details about these datasets are summarized in Table 1. 

\begin{table}[thbp]
  \caption{Detailed dataset descriptions}
  \label{tab:dataset}
  \centering
  \begin{threeparttable}
  \resizebox{\columnwidth}{!}
	{  
  \begin{tabular}{c|c|c|c|c}
    \toprule
    Dataset & Variants  & Dataset Size & Frequency & Domain \\
    \toprule
     \update{ETTm1} & 7 & (34465, 11521, 11521) & 15min & Electricity\\
    \midrule
    Weather & 21 & (36792, 5271, 10540) & 10min & Weather\\
    \midrule
    ECL & 321  & (18317, 2633, 5261) & Hourly & Electricity \\
    \midrule
    Solar-Energy & 137  & (36601, 5161, 10417) & 10min & Energy \\
    \midrule
    \update{PEMS03} & 358  & (15617, 5135, 5135) & 5min & Trffic\\
    \midrule
    \update{PEMS04} & 307  & (10172, 3375, 3375) & 5min & Trffic\\
    \midrule
    \update{PEMS07} & 883  & (16911, 5622, 5622) & 5min & Trffic\\
    \midrule
    \update{PEMS08} & 170  & (10690, 3548, 3548) & 5min & Trffic\\
    \bottomrule
    \end{tabular}
    }
  \end{threeparttable}
\end{table}

The dataset specifications are as follows:
(1) Four subsets from the PEMS dataset (PEMS03, PEMS04, PEMS07, PEMS08).
The PEMS dataset comprises spatiotemporal series data collected from California's public transportation network, widely used for traffic flow prediction and spatiotemporal sequence analysis. It records traffic flow, speed, and occupancy at 5-minute intervals.
(2) The Weather dataset.
Maintained by the Max Planck Institute for Biogeochemistry, this dataset is commonly employed in meteorological forecasting and climate analysis research. It contains 21 meteorological variables (including temperature, humidity, wind speed, and precipitation) recorded at 10-minute intervals throughout 2020.
(3) The ECL dataset.
This dataset documents hourly electricity consumption from 321 customers between 2012-2014, primarily used for power load forecasting and energy management studies.
(4) The Solar-Energy dataset.
Recording 10-minute interval power generation from 137 photovoltaic plants during 2006, this dataset is typically applied in solar power prediction and renewable energy management research.
(5) The ETTm1 dataset.
A subset of the Electricity Transformer Temperature (ETT) dataset provided by China's State Grid, ETTm1 is extensively used for power equipment condition monitoring and fault prediction. It contains 15-minute interval load and oil temperature measurements from July 2016 to July 2018.
All datasets were split into training, validation, and test sets following standard protocols. 

2) \textbf{Baselines.} We evaluated our models against 11 representative forecasting baselines, covering both channel-independent and channel-mixing SOTA models. The baselines include: S-Mamba \cite{wang2025S-Dmamba}, iTransformer \cite{liu2024itransformer}, RLinear \cite{li2023revisiting}, PatchTST \cite{Yuqietal-2023-PatchTST}, Crossformer \cite{zhang2022crossformer}, TiDE \cite{1-TiDE},TimesNet \cite{wu2022timesnet}, DLinear \cite{3-DLinear}, SCINet \cite{4-SCINet}, FEDformer \cite{zhou2022fedformer}, Stationary \cite{NEURIPS2022_4054556f}.

\subsection{Experimental Details} 
All computational experiments are implemented in PyTorch 3.10.9 and executed on an NVIDIA GeForce RTX 3090 GPU, maintaining deterministic computation through CUDA 11.4.  During model training, we employ the L2 loss function and the ADAM optimizer for iterative parameter updates. The training procedure is configured for 10 epochs, with early stopping incorporated when necessary to ensure training efficiency and effectiveness. The specific experimental setup and parameter configurations are presented in tabular form in the appendix. The formal procedural steps of DC-Mamber's prediction phase are algorithmically detailed in Algorithm \ref{The Forecasting Procedure of DC-Mamber}.

To evaluate the performance of the proposed model, we use two widely recognized metrics: mean squared error (MSE) and mean absolute error (MAE). These metrics are essential for assessing the accuracy and robustness of the model's predictions in regression tasks. MSE is defined as the average of the squared differences between the predicted and actual values:
\begin{equation}
\text{MSE} = \frac{1}{n} \sum_{i=1}^{n} (y_i - \hat{y}_i)^2.
\end{equation}
where ${n}$ is the number of samples, ${y_i}$ is the true value, and $\hat{y}_i$ is the predicted value. MSE is sensitive to large errors, as squaring the differences amplifies the impact of larger deviations. This makes it suitable for tasks where precise predictions are needed. In contrast, MAE is the average of the absolute differences between the predicted and actual values:
\begin{equation}
\text{MAE} = \frac{1}{n} \sum_{i=1}^{n} |y_i - \hat{y}_i|.
\end{equation}
MAE provides a more robust measure of model performance by treating all errors equally, without applying additional weight to larger deviations. This makes the MAE less sensitive to outliers, as compared to the MSE, and therefore a better choice when the data contains noise or extreme values that could distort model evaluation.

\renewcommand{\algorithmicrequire}{\textbf{Input:}}
\renewcommand{\algorithmicensure}{\textbf{Output:}}

\begin{algorithm}[h]
\renewcommand{\arraystretch}{1.5} 
\caption{The Forecasting Procedure of DC-Mamber}
\label{The Forecasting Procedure of DC-Mamber}
\begin{algorithmic}[1]
\Require $\mathbf{X} =\left\{\mathbf{x}_{1}, \ldots, \mathbf{x}_{L}\right\} \in \mathbb{R}^{B\times L\times V}$
\Ensure $\mathbf{Y} =\left\{\mathbf{x}_{L+1}, \ldots, \mathbf{x}_{L+W}\right\} \in \mathbb{R}^{B\times W\times V}$

\State $X_{\mathrm{norm}}\in \mathbb{R}^{B\times L\times V}\leftarrow\mathrm{Norm}(X) $

\State $X_{tem}\in \mathbb{R}^{B\times L\times D}\leftarrow\text{T-Embedding}(X_{\mathrm{norm}})$

\State $X_{var}\in \mathbb{R}^{B\times V\times D}\leftarrow\text{V-Embedding}(X_{\mathrm{norm}})$

\State $\textbf{for i in DC-Mamber Layers do:}$

\State \hspace{1.5em}$Y^{\prime} \in \mathbb{R}^{B\times L\times D}\leftarrow\text{T-Encoder Layer}(X_{i})$\{$X_i$ represents the $i_\text{th}$ T-Encoder Layer's input\}

\State \hspace{1.5em}$Y^{\prime\prime}\in \mathbb{R}^{B\times V\times D}\leftarrow\text{V-Encoder Layer}(X_{i}^{\prime\prime})$\{$X_i''$ represents the $i_\text{th}$ V-Encoder Layer's input\}

\State $Y^{\prime}\in \mathbb{R}^{B\times D\times L}\leftarrow \mathrm{Permute(Y^{\prime})}$
\State $Y^{\prime}\in \mathbb{R}^{B\times D\times V}\leftarrow \mathrm{Linear(Y^{\prime})}$\{Linear Projection\}
\State $Y^{\prime}\in \mathbb{R}^{B\times V\times D}\leftarrow \mathrm{Permute(Y^{\prime})}$

\State ${Y}\in \mathbb{R}^{B\times V\times D}\leftarrow\text{Feature-fusion}({Y^{\prime}},Y^{\prime\prime} )$

\State $Y\in \mathbb{R}^{B\times V\times W}\leftarrow\mathrm{Projector(Y)}$

\State $Y\in \mathbb{R}^{B\times W\times V}\leftarrow \mathrm{Permute(Y)}$

\end{algorithmic}
\end{algorithm}

\subsection{Results and Analysis}  
In the systematic evaluation of MTSF tasks, Table \ref{full_baseline_results} compares the forecasting performance of DC-Mamber with 11 baseline models across 8 real-world datasets. For clear comparison, the best results are highlighted in red, while the second-best results are highlighted in blue and underlined. The experimental setup follows the standard configuration of the iTransformer framework and includes the following components: 1) The input sequence length is uniformly fixed at 96 time steps; 2) For the ECL, Solar-Energy, Weather, and ETTm1 datasets, we compare four representative forecast lengths: {96, 192, 336, 720}. For the four PEMS datasets (PEMS03, PEMS04, PEMS07, and PEMS08), we evaluate forecast lengths of {12, 24, 48, 96}; 3) The average of the four forecast length results (denoted as Avg) is reported as an indicator of overall model performance. It is important to note that, except for the models marked with an asterisk (*), which were reproduced in this study based on official code, other baseline comparison data were directly sourced from the published results in the original iTransformer paper. 

\begin{table*}[htbp]
	\caption{Multivariate forecasting results of DC-Mamber. We conducted a comparative analysis of the performance of existing baseline models across different forecast horizons. 
	}\label{full_baseline_results}

	\centering
	\resizebox{\textwidth}{!}
	{
		\begin{threeparttable}
				\renewcommand{\multirowsetup}{\centering}
				\setlength{\tabcolsep}{1pt}
				\begin{tabular}{c|c|cc|cc|cc|cc|cc|cc|cc|cc|cc|cc|cc|cc}
					\toprule
					\multicolumn{2}{c|}{\multirow{1}{*}{Models}} &
					\multicolumn{2}{c}{\rotatebox{0}{\scalebox{0.85}{\textbf{Ours}}}} & 
					\multicolumn{2}{c}{\rotatebox{0}{\scalebox{0.85}{S-Mamba* }}} & 
					\multicolumn{2}{c}{\rotatebox{0}{\scalebox{0.85}{iTransformer }}} &
					\multicolumn{2}{c}{\rotatebox{0}{\scalebox{0.85}{\update{RLinear}}}} &
					\multicolumn{2}{c}{\rotatebox{0}{\scalebox{0.85}{PatchTST}}} &
					\multicolumn{2}{c}{\rotatebox{0}{\scalebox{0.85}{Crossformer}}} &
					\multicolumn{2}{c}{\rotatebox{0}{\scalebox{0.85}{TiDE}}} &
					\multicolumn{2}{c}{\rotatebox{0}{\scalebox{0.85}{{TimesNet}}}} &
					\multicolumn{2}{c}{\rotatebox{0}{\scalebox{0.85}{DLinear}}} &
					\multicolumn{2}{c}{\rotatebox{0}{\scalebox{0.85}{SCINet}}} &
					\multicolumn{2}{c}{\rotatebox{0}{\scalebox{0.85}{FEDformer}}} &
					\multicolumn{2}{c}{\rotatebox{0}{\scalebox{0.85}{Stationary}}}  \\

					\cmidrule(lr){3-4} \cmidrule(lr){5-6}\cmidrule(lr){7-8} \cmidrule(lr){9-10}\cmidrule(lr){11-12}\cmidrule(lr){13-14} \cmidrule(lr){15-16} \cmidrule(lr){17-18} \cmidrule(lr){19-20} \cmidrule(lr){21-22} \cmidrule(lr){23-24} \cmidrule(lr){25-26}
					\multicolumn{2}{c|}{Metric}  & \scalebox{0.78}{MSE} & \scalebox{0.78}{MAE}  & \scalebox{0.78}{MSE} & \scalebox{0.78}{MAE}  & \scalebox{0.78}{MSE} & \scalebox{0.78}{MAE}  & \scalebox{0.78}{MSE} & \scalebox{0.78}{MAE}  & \scalebox{0.78}{MSE} & \scalebox{0.78}{MAE}  & \scalebox{0.78}{MSE} & \scalebox{0.78}{MAE} & \scalebox{0.78}{MSE} & \scalebox{0.78}{MAE} & \scalebox{0.78}{MSE} & \scalebox{0.78}{MAE} & \scalebox{0.78}{MSE} & \scalebox{0.78}{MAE} & \scalebox{0.78}{MSE} & \scalebox{0.78}{MAE} & \scalebox{0.78}{MSE} & \scalebox{0.78}{MAE} & \scalebox{0.78}{MSE} & \scalebox{0.78}{MAE} \\
					\toprule
					
					\multirow{5}{*}{\rotatebox{90}{\scalebox{0.95}{PEMS03}}}
					&  \scalebox{0.78}{12} & \boldres{\scalebox{0.78}{0.061}} &\boldres{\scalebox{0.78}{0.163}} &\secondres{\scalebox{0.78}{0.066}} &\secondres{\scalebox{0.78}{0.170}} & {\scalebox{0.78}{0.071}} &{\scalebox{0.78}{0.174}} &\scalebox{0.78}{0.126} &\scalebox{0.78}{0.236} &\scalebox{0.78}{0.099} &\scalebox{0.78}{0.216} &\scalebox{0.78}{0.090} &\scalebox{0.78}{0.203} & \scalebox{0.78}{0.178} & \scalebox{0.78}{0.305} &\scalebox{0.78}{0.085} &\scalebox{0.78}{0.192} &\scalebox{0.78}{0.122} &\scalebox{0.78}{0.243} &\secondres{\scalebox{0.78}{0.066}} &{\scalebox{0.78}{0.172}} &\scalebox{0.78}{0.126} &\scalebox{0.78}{0.251} &\scalebox{0.78}{0.081} &\scalebox{0.78}{0.188}     \\
     
					& \scalebox{0.78}{24} & \boldres{\scalebox{0.78}{0.080}} &\boldres{\scalebox{0.78}{0.186}} &{\scalebox{0.78}{0.088}} &\secondres{\scalebox{0.78}{0.197}} &{\scalebox{0.78}{0.093}} &{\scalebox{0.78}{0.201}} &\scalebox{0.78}{0.246} &\scalebox{0.78}{0.334} &\scalebox{0.78}{0.142} &\scalebox{0.78}{0.259} &\scalebox{0.78}{0.121} &\scalebox{0.78}{0.240} & \scalebox{0.78}{0.257} & \scalebox{0.78}{0.371} &\scalebox{0.78}{0.118} &\scalebox{0.78}{0.223} &\scalebox{0.78}{0.201} &\scalebox{0.78}{0.317} &\secondres{\scalebox{0.78}{0.085}} &{\scalebox{0.78}{0.198}} &\scalebox{0.78}{0.149} &\scalebox{0.78}{0.275} &\scalebox{0.78}{0.105} &\scalebox{0.78}{0.214}  \\
     
					& \scalebox{0.78}{48} & \boldres{\scalebox{0.78}{0.114}} &\boldres{\scalebox{0.78}{0.225}} &{\scalebox{0.78}{0.165}} &{\scalebox{0.78}{0.277}} &\secondres{\scalebox{0.78}{0.125}} &\secondres{\scalebox{0.78}{0.236}} &\scalebox{0.78}{0.551} &\scalebox{0.78}{0.529} &\scalebox{0.78}{0.211} &\scalebox{0.78}{0.319}  &\scalebox{0.78}{0.202} &\scalebox{0.78}{0.317} & \scalebox{0.78}{0.379}& \scalebox{0.78}{0.463} &\scalebox{0.78}{0.155} &\scalebox{0.78}{0.260} &\scalebox{0.78}{0.333} &\scalebox{0.78}{0.425} &{\scalebox{0.78}{0.127}} &{\scalebox{0.78}{0.238}} &\scalebox{0.78}{0.227} &\scalebox{0.78}{0.348} &\scalebox{0.78}{0.154} &\scalebox{0.78}{0.257}  \\
     
					& \scalebox{0.78}{96} & \secondres{\scalebox{0.78}{0.169}} &\secondres{\scalebox{0.78}{0.280}} &{\scalebox{0.78}{0.226}} &{\scalebox{0.78}{0.321}} &\boldres{\scalebox{0.78}{0.164}} &\boldres{\scalebox{0.78}{0.275}} &\scalebox{0.78}{1.057} &\scalebox{0.78}{0.787} &\scalebox{0.78}{0.269} &\scalebox{0.78}{0.370} &\scalebox{0.78}{0.262} &\scalebox{0.78}{0.367} & \scalebox{0.78}{0.490}& \scalebox{0.78}{0.539} &\scalebox{0.78}{0.228} &\scalebox{0.78}{0.317} &\scalebox{0.78}{0.457} &\scalebox{0.78}{0.515} &{\scalebox{0.78}{0.178}} &{\scalebox{0.78}{0.287}} &\scalebox{0.78}{0.348} &\scalebox{0.78}{0.434} &\scalebox{0.78}{0.247} &\scalebox{0.78}{0.336} \\
     
					\cmidrule(lr){2-24}
					& \scalebox{0.78}{Avg} & \boldres{\scalebox{0.78}{0.106}} &\boldres{\scalebox{0.78}{0.214}} &{\scalebox{0.78}{0.136}} &{\scalebox{0.78}{0.241}} &\secondres{\scalebox{0.78}{0.113}} &\secondres{\scalebox{0.78}{0.221}} &\scalebox{0.78}{0.495} &\scalebox{0.78}{0.472} &\scalebox{0.78}{0.180} &\scalebox{0.78}{0.291} &\scalebox{0.78}{0.169} &\scalebox{0.78}{0.281} & \scalebox{0.78}{0.326}& \scalebox{0.78}{0.419} &\scalebox{0.78}{0.147} &\scalebox{0.78}{0.248} &\scalebox{0.78}{0.278} &\scalebox{0.78}{0.375} &{\scalebox{0.78}{0.114}} &{\scalebox{0.78}{0.224}} &\scalebox{0.78}{0.213} &\scalebox{0.78}{0.327} &\scalebox{0.78}{0.147} &\scalebox{0.78}{0.249} \\
					
					\midrule
					\multirow{5}{*}{\update{\rotatebox{90}{\scalebox{0.95}{PEMS04}}}} 
					&  \scalebox{0.78}{12} & \boldres{\scalebox{0.78}{0.069}} &\boldres{\scalebox{0.78}{0.163}} &\secondres{\scalebox{0.78}{0.072}} &\secondres{\scalebox{0.78}{0.177}} &{\scalebox{0.78}{0.078}} &{\scalebox{0.78}{0.183}} &\scalebox{0.78}{0.138} &\scalebox{0.78}{0.252} &\scalebox{0.78}{0.105} &\scalebox{0.78}{0.224} &\scalebox{0.78}{0.098} &\scalebox{0.78}{0.218} & \scalebox{0.78}{0.219}& \scalebox{0.78}{0.340} &\scalebox{0.78}{0.087} &\scalebox{0.78}{0.195} &\scalebox{0.78}{0.148} &\scalebox{0.78}{0.272} &{\scalebox{0.78}{0.073}} &\secondres{\scalebox{0.78}{0.177}} &\scalebox{0.78}{0.138} &\scalebox{0.78}{0.262} &\scalebox{0.78}{0.088} &\scalebox{0.78}{0.196}  \\
     
					& \scalebox{0.78}{24} & \boldres{\scalebox{0.78}{0.076}} &\boldres{\scalebox{0.78}{0.179}} &\secondres{\scalebox{0.78}{0.084}} &\secondres{\scalebox{0.78}{0.192}} &{\scalebox{0.78}{0.095}} &{\scalebox{0.78}{0.205}} &\scalebox{0.78}{0.258} &\scalebox{0.78}{0.348} &\scalebox{0.78}{0.153} &\scalebox{0.78}{0.275} &\scalebox{0.78}{0.131} &\scalebox{0.78}{0.256} & \scalebox{0.78}{0.292}& \scalebox{0.78}{0.398} &\scalebox{0.78}{0.103} &\scalebox{0.78}{0.215} &\scalebox{0.78}{0.224} &\scalebox{0.78}{0.340} &\secondres{\scalebox{0.78}{0.084}} &{\scalebox{0.78}{0.193}} &\scalebox{0.78}{0.177} &\scalebox{0.78}{0.293} &\scalebox{0.78}{0.104} &\scalebox{0.78}{0.216} \\
     
					& \scalebox{0.78}{48} & \boldres{\scalebox{0.78}{0.088}} &\boldres{\scalebox{0.78}{0.195}} &{\scalebox{0.78}{0.101}} &{\scalebox{0.78}{0.212}} &{\scalebox{0.78}{0.120}} &{\scalebox{0.78}{0.233}} &\scalebox{0.78}{0.572} &\scalebox{0.78}{0.544} &\scalebox{0.78}{0.229} &\scalebox{0.78}{0.339} &\scalebox{0.78}{0.205} &\scalebox{0.78}{0.326} & \scalebox{0.78}{0.409}& \scalebox{0.78}{0.478} &\scalebox{0.78}{0.136} &\scalebox{0.78}{0.250} &\scalebox{0.78}{0.355} &\scalebox{0.78}{0.437} &\secondres{\scalebox{0.78}{0.099}} &\secondres{\scalebox{0.78}{0.211}} &\scalebox{0.78}{0.270} &\scalebox{0.78}{0.368} &\scalebox{0.78}{0.137} &\scalebox{0.78}{0.251}  \\
     
					& \scalebox{0.78}{96} & \boldres{\scalebox{0.78}{0.100}} &\boldres{\scalebox{0.78}{0.207}} &{\scalebox{0.78}{0.127}} &{\scalebox{0.78}{0.236}} &{\scalebox{0.78}{0.150}} &{\scalebox{0.78}{0.262}} &\scalebox{0.78}{1.137} &\scalebox{0.78}{0.820} &\scalebox{0.78}{0.291} &\scalebox{0.78}{0.389} &\scalebox{0.78}{0.402} &\scalebox{0.78}{0.457} & \scalebox{0.78}{0.492}& \scalebox{0.78}{0.532} &\scalebox{0.78}{0.190} &\scalebox{0.78}{0.303} &\scalebox{0.78}{0.452} &\scalebox{0.78}{0.504} &\secondres{\scalebox{0.78}{0.114}} &\secondres{\scalebox{0.78}{0.227}} &\scalebox{0.78}{0.341} &\scalebox{0.78}{0.427} &\scalebox{0.78}{0.186} &\scalebox{0.78}{0.297}  \\
     
					\cmidrule(lr){2-24}
					& \scalebox{0.78}{Avg} & \boldres{\scalebox{0.78}{0.083}} &\boldres{\scalebox{0.78}{0.188}} &{\scalebox{0.78}{0.096}} &{\scalebox{0.78}{0.204}} &{\scalebox{0.78}{0.111}} &{\scalebox{0.78}{0.221}} &\scalebox{0.78}{0.526} &\scalebox{0.78}{0.491} &\scalebox{0.78}{0.195} &\scalebox{0.78}{0.307} &\scalebox{0.78}{0.209} &\scalebox{0.78}{0.314} & \scalebox{0.78}{0.353}& \scalebox{0.78}{0.437} &\scalebox{0.78}{0.129} &\scalebox{0.78}{0.241} &\scalebox{0.78}{0.295} &\scalebox{0.78}{0.388} &\secondres{\scalebox{0.78}{0.092}} &\secondres{\scalebox{0.78}{0.202}} &\scalebox{0.78}{0.231} &\scalebox{0.78}{0.337} &\scalebox{0.78}{0.127} &\scalebox{0.78}{0.240} \\
					
					\midrule
					\multirow{5}{*}{\update{\rotatebox{90}{\scalebox{0.95}{PEMS07}}}}
					&  \scalebox{0.78}{12} & \boldres{\scalebox{0.78}{0.059}} &\boldres{\scalebox{0.78}{0.151}} &\secondres{\scalebox{0.78}{0.060}} &\secondres{\scalebox{0.78}{0.157}} &{\scalebox{0.78}{0.067}} &{\scalebox{0.78}{0.165}} &\scalebox{0.78}{0.118} &\scalebox{0.78}{0.235} &\scalebox{0.78}{0.095} &\scalebox{0.78}{0.207} &\scalebox{0.78}{0.094} &\scalebox{0.78}{0.200} & \scalebox{0.78}{0.173}& \scalebox{0.78}{0.304} &\scalebox{0.78}{0.082} &\scalebox{0.78}{0.181} &\scalebox{0.78}{0.115} &\scalebox{0.78}{0.242} &{\scalebox{0.78}{0.068}} &{\scalebox{0.78}{0.171}} &\scalebox{0.78}{0.109} &\scalebox{0.78}{0.225} &\scalebox{0.78}{0.083} &\scalebox{0.78}{0.185}  \\
     
					& \scalebox{0.78}{24} & \boldres{\scalebox{0.78}{0.071}} &\boldres{\scalebox{0.78}{0.168}} &\secondres{\scalebox{0.78}{0.077}} &\secondres{\scalebox{0.78}{0.178}} &{\scalebox{0.78}{0.088}} &{\scalebox{0.78}{0.190}} &\scalebox{0.78}{0.242} &\scalebox{0.78}{0.341} &\scalebox{0.78}{0.150} &\scalebox{0.78}{0.262} &\scalebox{0.78}{0.139} &\scalebox{0.78}{0.247} & \scalebox{0.78}{0.271}& \scalebox{0.78}{0.383} &\scalebox{0.78}{0.101} &\scalebox{0.78}{0.204} &\scalebox{0.78}{0.210} &\scalebox{0.78}{0.329} &{\scalebox{0.78}{0.119}} &{\scalebox{0.78}{0.225}} &\scalebox{0.78}{0.125} &\scalebox{0.78}{0.244} &\scalebox{0.78}{0.102} &\scalebox{0.78}{0.207} \\
     
					& \scalebox{0.78}{48} & \boldres{\scalebox{0.78}{0.090}} &\boldres{\scalebox{0.78}{0.188}} &\secondres{\scalebox{0.78}{0.095}} &\secondres{\scalebox{0.78}{0.197}} &{\scalebox{0.78}{0.110}} &{\scalebox{0.78}{0.215}} &\scalebox{0.78}{0.562} &\scalebox{0.78}{0.541} &\scalebox{0.78}{0.253} &\scalebox{0.78}{0.340} &\scalebox{0.78}{0.311} &\scalebox{0.78}{0.369} & \scalebox{0.78}{0.446}& \scalebox{0.78}{0.495} &\scalebox{0.78}{0.134} &\scalebox{0.78}{0.238} &\scalebox{0.78}{0.398} &\scalebox{0.78}{0.458} &{\scalebox{0.78}{0.149}} &{\scalebox{0.78}{0.237}} &\scalebox{0.78}{0.165} &\scalebox{0.78}{0.288} &\scalebox{0.78}{0.136} &\scalebox{0.78}{0.240} \\
     
					& \scalebox{0.78}{96} & \boldres{\scalebox{0.78}{0.103}} &\boldres{\scalebox{0.78}{0.200}} &\secondres{\scalebox{0.78}{0.118}} &\secondres{\scalebox{0.78}{0.218}} &{\scalebox{0.78}{0.139}} &{\scalebox{0.78}{0.245}} &\scalebox{0.78}{1.096} &\scalebox{0.78}{0.795} &\scalebox{0.78}{0.346} &\scalebox{0.78}{0.404} &\scalebox{0.78}{0.396} &\scalebox{0.78}{0.442} & \scalebox{0.78}{0.628}& \scalebox{0.78}{0.577} &\scalebox{0.78}{0.181} &\scalebox{0.78}{0.279} &\scalebox{0.78}{0.594} &\scalebox{0.78}{0.553} &{\scalebox{0.78}{0.141}} &{\scalebox{0.78}{0.234}} &\scalebox{0.78}{0.262} &\scalebox{0.78}{0.376} &\scalebox{0.78}{0.187} &\scalebox{0.78}{0.287} \\
     
					\cmidrule(lr){2-24}
					& \scalebox{0.78}{Avg} & \boldres{\scalebox{0.78}{0.081}} &\boldres{\scalebox{0.78}{0.177}} &\secondres{\scalebox{0.78}{0.088}} &\secondres{\scalebox{0.78}{0.188}} &{\scalebox{0.78}{0.101}} &{\scalebox{0.78}{0.204}} &\scalebox{0.78}{0.504} &\scalebox{0.78}{0.478} &\scalebox{0.78}{0.211} &\scalebox{0.78}{0.303} &\scalebox{0.78}{0.235} &\scalebox{0.78}{0.315} & \scalebox{0.78}{0.380}& \scalebox{0.78}{0.440} &\scalebox{0.78}{0.124} &\scalebox{0.78}{0.225} &\scalebox{0.78}{0.329} &\scalebox{0.78}{0.395} &{\scalebox{0.78}{0.119}} &{\scalebox{0.78}{0.234}} &\scalebox{0.78}{0.165} &\scalebox{0.78}{0.283} &\scalebox{0.78}{0.127} &\scalebox{0.78}{0.230} \\
					
					\midrule
					\multirow{5}{*}{\update{\rotatebox{90}{\scalebox{0.95}{PEMS08}}}}
					&  \scalebox{0.78}{12} & \boldres{\scalebox{0.78}{0.076}} &\boldres{\scalebox{0.78}{0.172}} &\boldres{\scalebox{0.78}{0.076}} &\secondres{\scalebox{0.78}{0.178}} &{\scalebox{0.78}{0.079}} &{\scalebox{0.78}{0.182}} &\scalebox{0.78}{0.133} &\scalebox{0.78}{0.247} &\scalebox{0.78}{0.168} &\scalebox{0.78}{0.232} &\scalebox{0.78}{0.165} &\scalebox{0.78}{0.214} & \scalebox{0.78}{0.227}& \scalebox{0.78}{0.343} &\scalebox{0.78}{0.112} &\scalebox{0.78}{0.212} &\scalebox{0.78}{0.154} &\scalebox{0.78}{0.276} &{\scalebox{0.78}{0.087}} &{\scalebox{0.78}{0.184}} &\scalebox{0.78}{0.173} &\scalebox{0.78}{0.273} &\scalebox{0.78}{0.109} &\scalebox{0.78}{0.207} \\
     
					& \scalebox{0.78}{24} & \boldres{\scalebox{0.78}{0.100}} &\boldres{\scalebox{0.78}{0.198}} &\secondres{\scalebox{0.78}{0.110}} &\secondres{\scalebox{0.78}{0.216}} &{\scalebox{0.78}{0.115}} &{\scalebox{0.78}{0.219}} &\scalebox{0.78}{0.249} &\scalebox{0.78}{0.343} &\scalebox{0.78}{0.224} &\scalebox{0.78}{0.281} &\scalebox{0.78}{0.215} &\scalebox{0.78}{0.260} & \scalebox{0.78}{0.318}& \scalebox{0.78}{0.409} &\scalebox{0.78}{0.141} &\scalebox{0.78}{0.238} &\scalebox{0.78}{0.248} &\scalebox{0.78}{0.353} &{\scalebox{0.78}{0.122}} &{\scalebox{0.78}{0.221}} &\scalebox{0.78}{0.210} &\scalebox{0.78}{0.301} &\scalebox{0.78}{0.140} &\scalebox{0.78}{0.236} \\
     
					& \scalebox{0.78}{48} & {\scalebox{0.78}{0.196}} &\boldres{\scalebox{0.78}{0.223}} &\boldres{\scalebox{0.78}{0.165}} &{\scalebox{0.78}{0.252}} &\secondres{\scalebox{0.78}{0.186}} &\secondres{\scalebox{0.78}{0.235}} &\scalebox{0.78}{0.569} &\scalebox{0.78}{0.544} &\scalebox{0.78}{0.321} &\scalebox{0.78}{0.354} &\scalebox{0.78}{0.315} &\scalebox{0.78}{0.355} & \scalebox{0.78}{0.497}& \scalebox{0.78}{0.510} &\scalebox{0.78}{0.198} &\scalebox{0.78}{0.283} &\scalebox{0.78}{0.440} &\scalebox{0.78}{0.470} &{\scalebox{0.78}{0.189}} &x{\scalebox{0.78}{0.270}} &\scalebox{0.78}{0.320} &\scalebox{0.78}{0.394} &\scalebox{0.78}{0.211} &\scalebox{0.78}{0.294} \\
     
					& \scalebox{0.78}{96} & \boldres{\scalebox{0.78}{0.217}} &\boldres{\scalebox{0.78}{0.244}} &{\scalebox{0.78}{0.274}} &{\scalebox{0.78}{0.327}} &\secondres{\scalebox{0.78}{0.221}} &\secondres{\scalebox{0.78}{0.267}} &\scalebox{0.78}{1.166} &\scalebox{0.78}{0.814} &\scalebox{0.78}{0.408} &\scalebox{0.78}{0.417} &\scalebox{0.78}{0.377} &\scalebox{0.78}{0.397} & \scalebox{0.78}{0.721}& \scalebox{0.78}{0.592} &\scalebox{0.78}{0.320} &\scalebox{0.78}{0.351} &\scalebox{0.78}{0.674} &\scalebox{0.78}{0.565} &{\scalebox{0.78}{0.236}} &{\scalebox{0.78}{0.300}} &\scalebox{0.78}{0.442} &\scalebox{0.78}{0.465} &\scalebox{0.78}{0.345} &\scalebox{0.78}{0.367}  \\
     
					\cmidrule(lr){2-24}
					& \scalebox{0.78}{Avg} & \boldres{\scalebox{0.78}{0.147}} &\boldres{\scalebox{0.78}{0.209}} &{\scalebox{0.78}{0.156}} &{\scalebox{0.78}{0.243}} &\secondres{\scalebox{0.78}{0.150}} &\secondres{\scalebox{0.78}{0.226}} &\scalebox{0.78}{0.529} &\scalebox{0.78}{0.487} &\scalebox{0.78}{0.280} &\scalebox{0.78}{0.321} &\scalebox{0.78}{0.268} &\scalebox{0.78}{0.307} & \scalebox{0.78}{0.441}& \scalebox{0.78}{0.464} &\scalebox{0.78}{0.193} &\scalebox{0.78}{0.271} &\scalebox{0.78}{0.379} &\scalebox{0.78}{0.416} &{\scalebox{0.78}{0.158}} &{\scalebox{0.78}{0.244}} &\scalebox{0.78}{0.286} &\scalebox{0.78}{0.358} &\scalebox{0.78}{0.201} &\scalebox{0.78}{0.276}  \\
     
					\midrule
                    \multirow{5}{*}{\rotatebox{90}{\scalebox{0.95}{Electricity}}} 
					&  \scalebox{0.78}{96} & \boldres{\scalebox{0.78}{0.139}} &\boldres{\scalebox{0.78}{0.235}} &\boldres{\scalebox{0.78}{0.139}} &\boldres{\scalebox{0.78}{0.235}} & {\scalebox{0.78}{0.148}} & {\scalebox{0.78}{0.240}} & \scalebox{0.78}{0.201} & \scalebox{0.78}{0.281} & \scalebox{0.78}{0.195} & \scalebox{0.78}{0.285} & \scalebox{0.78}{0.219} & \scalebox{0.78}{0.314} & \scalebox{0.78}{0.237} & \scalebox{0.78}{0.329} &{\scalebox{0.78}{0.168}} &{\scalebox{0.78}{0.272}} &\scalebox{0.78}{0.197} &\scalebox{0.78}{0.282} & \scalebox{0.78}{0.247} & \scalebox{0.78}{0.345} &\scalebox{0.78}{0.193} &\scalebox{0.78}{0.308} &{\scalebox{0.78}{0.169}} &{\scalebox{0.78}{0.273}}  \\ 
     
					& \scalebox{0.78}{192}  & {\scalebox{0.78}{0.163}} &{\scalebox{0.78}{0.259}} &\boldres{\scalebox{0.78}{0.161}} &\secondres{\scalebox{0.78}{0.258}} & \secondres{\scalebox{0.78}{0.162}} & \boldres{\scalebox{0.78}{0.253}} & \scalebox{0.78}{0.201} & {\scalebox{0.78}{0.283}} & \scalebox{0.78}{0.199} & \scalebox{0.78}{0.289} & \scalebox{0.78}{0.231} & \scalebox{0.78}{0.322} & \scalebox{0.78}{0.236} & \scalebox{0.78}{0.330} &{\scalebox{0.78}{0.184}} &\scalebox{0.78}{0.289} &\scalebox{0.78}{0.196} &{\scalebox{0.78}{0.285}} & \scalebox{0.78}{0.257} & \scalebox{0.78}{0.355} &\scalebox{0.78}{0.201} &\scalebox{0.78}{0.315} &{\scalebox{0.78}{0.182}} &\scalebox{0.78}{0.286} \\ 
     
					& \scalebox{0.78}{336}  & \boldres{\scalebox{0.78}{0.176}} &\secondres{\scalebox{0.78}{0.273}} &{\scalebox{0.78}{0.181}} &{\scalebox{0.78}{0.278}} & \secondres{\scalebox{0.78}{0.178}} & \boldres{\scalebox{0.78}{0.269}} & \scalebox{0.78}{0.215} & {\scalebox{0.78}{0.298}} & \scalebox{0.78}{0.215} & \scalebox{0.78}{0.305} & \scalebox{0.78}{0.246} & \scalebox{0.78}{0.337} & \scalebox{0.78}{0.249} & \scalebox{0.78}{0.344} &{\scalebox{0.78}{0.198}} &{\scalebox{0.78}{0.300}} &\scalebox{0.78}{0.209} &{\scalebox{0.78}{0.301}} & \scalebox{0.78}{0.269} & \scalebox{0.78}{0.369} &\scalebox{0.78}{0.214} &\scalebox{0.78}{0.329} &{\scalebox{0.78}{0.200}} &\scalebox{0.78}{0.304} \\ 
     
					& \scalebox{0.78}{720}  & \boldres{\scalebox{0.78}{0.197}} &\boldres{\scalebox{0.78}{0.294}} &\secondres{\scalebox{0.78}{0.201}} &\secondres{\scalebox{0.78}{0.298}} & {\scalebox{0.78}{0.225}} & {\scalebox{0.78}{0.317}} & \scalebox{0.78}{0.257} & \scalebox{0.78}{0.331} & \scalebox{0.78}{0.256} & \scalebox{0.78}{0.337} & \scalebox{0.78}{0.280} & \scalebox{0.78}{0.363} & \scalebox{0.78}{0.284} & \scalebox{0.78}{0.373} &{\scalebox{0.78}{0.220}} &{\scalebox{0.78}{0.320}} &\scalebox{0.78}{0.245} &\scalebox{0.78}{0.333} & \scalebox{0.78}{0.299} & \scalebox{0.78}{0.390} &\scalebox{0.78}{0.246} &\scalebox{0.78}{0.355} &{\scalebox{0.78}{0.222}} &{\scalebox{0.78}{0.321}}  \\ 
     
					\cmidrule(lr){2-24}
					& \scalebox{0.78}{Avg} & \boldres{\scalebox{0.78}{0.169}} &\boldres{\scalebox{0.78}{0.265}} &\secondres{\scalebox{0.78}{0.171}} &\secondres{\scalebox{0.78}{0.267}} & {\scalebox{0.78}{0.178}} & {\scalebox{0.78}{0.270}} & \scalebox{0.78}{0.219} & \scalebox{0.78}{0.298} & \scalebox{0.78}{0.216} & \scalebox{0.78}{0.304} & \scalebox{0.78}{0.244} & \scalebox{0.78}{0.334} & \scalebox{0.78}{0.251} & \scalebox{0.78}{0.344} &{\scalebox{0.78}{0.192}} &{\scalebox{0.78}{0.295}} &\scalebox{0.78}{0.212} &\scalebox{0.78}{0.300} & \scalebox{0.78}{0.268} & \scalebox{0.78}{0.365} &\scalebox{0.78}{0.214} &\scalebox{0.78}{0.327} &{\scalebox{0.78}{0.193}} &{\scalebox{0.78}{0.296}} \\ 
					\midrule

				\multirow{5}{*}{\rotatebox{90}{\scalebox{0.95}{Solar-Energy}}} 
					&  \scalebox{0.78}{96}  & \boldres{\scalebox{0.78}{0.200}} &\boldres{\scalebox{0.78}{0.228}} &{\scalebox{0.78}{0.208}} &{\scalebox{0.78}{0.246}} &\secondres{\scalebox{0.78}{0.203}} &\secondres{\scalebox{0.78}{0.237}} & \scalebox{0.78}{0.322} & \scalebox{0.78}{0.339} & {\scalebox{0.78}{0.234}} & {\scalebox{0.78}{0.286}} &\scalebox{0.78}{0.310} &\scalebox{0.78}{0.331} &\scalebox{0.78}{0.312} &\scalebox{0.78}{0.399} &\scalebox{0.78}{0.250} &\scalebox{0.78}{0.292} &\scalebox{0.78}{0.290} &\scalebox{0.78}{0.378} &\scalebox{0.78}{0.237} &\scalebox{0.78}{0.344} &\scalebox{0.78}{0.242} &\scalebox{0.78}{0.342} &\scalebox{0.78}{0.215} &\scalebox{0.78}{0.249} \\ 
     
					& \scalebox{0.78}{192} & \secondres{\scalebox{0.78}{0.235}} &\boldres{\scalebox{0.78}{0.261}} &{\scalebox{0.78}{0.240}} &\secondres{\scalebox{0.78}{0.272}} &\boldres{\scalebox{0.78}{0.233}} &\boldres{\scalebox{0.78}{0.261}} & \scalebox{0.78}{0.359} & \scalebox{0.78}{0.356}& {\scalebox{0.78}{0.267}} & {\scalebox{0.78}{0.310}} &\scalebox{0.78}{0.734} &\scalebox{0.78}{0.725} &\scalebox{0.78}{0.339} &\scalebox{0.78}{0.416} &\scalebox{0.78}{0.296} &\scalebox{0.78}{0.318} &\scalebox{0.78}{0.320} &\scalebox{0.78}{0.398} &\scalebox{0.78}{0.280} &\scalebox{0.78}{0.380} &\scalebox{0.78}{0.285} &\scalebox{0.78}{0.380} &\scalebox{0.78}{0.254} &\scalebox{0.78}{0.272} \\ 
     
					& \scalebox{0.78}{336}  & \boldres{\scalebox{0.78}{0.247}} &\boldres{\scalebox{0.78}{0.272}} &{\scalebox{0.78}{0.262}} &{\scalebox{0.78}{0.290}} &\secondres{\scalebox{0.78}{0.248}} &\secondres{\scalebox{0.78}{0.273}} & \scalebox{0.78}{0.397} & \scalebox{0.78}{0.369}& {\scalebox{0.78}{0.290}}  &{\scalebox{0.78}{0.315}} &\scalebox{0.78}{0.750} &\scalebox{0.78}{0.735} &\scalebox{0.78}{0.368} &\scalebox{0.78}{0.430} &\scalebox{0.78}{0.319} &\scalebox{0.78}{0.330} &\scalebox{0.78}{0.353} &\scalebox{0.78}{0.415} &\scalebox{0.78}{0.304} &\scalebox{0.78}{0.389} &\scalebox{0.78}{0.282} &\scalebox{0.78}{0.376} &\scalebox{0.78}{0.290} &\scalebox{0.78}{0.296} \\ 
     
					& \scalebox{0.78}{720}  & \boldres{\scalebox{0.78}{0.248}} &\boldres{\scalebox{0.78}{0.274}} &{\scalebox{0.78}{0.267}} &{\scalebox{0.78}{0.293}} &\secondres{\scalebox{0.78}{0.249}} &\secondres{\scalebox{0.78}{0.275}} & \scalebox{0.78}{0.397} & \scalebox{0.78}{0.356} & {\scalebox{0.78}{0.289}} &{\scalebox{0.78}{0.317}} &\scalebox{0.78}{0.769} &\scalebox{0.78}{0.765} &\scalebox{0.78}{0.370} &\scalebox{0.78}{0.425} &\scalebox{0.78}{0.338} &\scalebox{0.78}{0.337} &\scalebox{0.78}{0.356} &\scalebox{0.78}{0.413} &\scalebox{0.78}{0.308} &\scalebox{0.78}{0.388} &\scalebox{0.78}{0.357} &\scalebox{0.78}{0.427} &\scalebox{0.78}{0.285} &\scalebox{0.78}{0.295}  \\ 
     
					\cmidrule(lr){2-24}
					& \scalebox{0.78}{Avg}  & \boldres{\scalebox{0.78}{0.233}} &\boldres{\scalebox{0.78}{0.259}} &{\scalebox{0.78}{0.244}} &{\scalebox{0.78}{0.275}} &\boldres{\scalebox{0.78}{0.233}} &\secondres{\scalebox{0.78}{0.262}} & \scalebox{0.78}{0.369} & \scalebox{0.78}{0.356} &{\scalebox{0.78}{0.270}} &{\scalebox{0.78}{0.307}} &\scalebox{0.78}{0.641} &\scalebox{0.78}{0.639} &\scalebox{0.78}{0.347} &\scalebox{0.78}{0.417} &\scalebox{0.78}{0.301} &\scalebox{0.78}{0.319} &\scalebox{0.78}{0.330} &\scalebox{0.78}{0.401} &\scalebox{0.78}{0.282} &\scalebox{0.78}{0.375} &\scalebox{0.78}{0.291} &\scalebox{0.78}{0.381} &\scalebox{0.78}{0.261} &\scalebox{0.78}{0.381} \\ 

                \midrule
					
					\multirow{5}{*}{\rotatebox{90}{\scalebox{0.95}{Weather}}} 
					&  \scalebox{0.78}{96}  &\boldres{\scalebox{0.78}{0.158}} &\boldres{\scalebox{0.78}{0.206}} &{\scalebox{0.78}{0.165}} &\secondres{\scalebox{0.78}{0.209}} & \scalebox{0.78}{0.174} & {\scalebox{0.78}{0.214}} & \scalebox{0.78}{0.192} & \scalebox{0.78}{0.232} & \scalebox{0.78}{0.177} & {\scalebox{0.78}{0.218}} & \boldres{\scalebox{0.78}{0.158}} & \scalebox{0.78}{0.230}  & \scalebox{0.78}{0.202} & \scalebox{0.78}{0.261} &{\scalebox{0.78}{0.172}} &{\scalebox{0.78}{0.220}} & \scalebox{0.78}{0.196} &\scalebox{0.78}{0.255} & \scalebox{0.78}{0.221} & \scalebox{0.78}{0.306} & \scalebox{0.78}{0.217} &\scalebox{0.78}{0.296} & {\scalebox{0.78}{0.173}} &{\scalebox{0.78}{0.223}}  \\ 
     
					& \scalebox{0.78}{192}  & {\scalebox{0.78}{0.218}} &{\scalebox{0.78}{0.260}} &\secondres{\scalebox{0.78}{0.215}} &\secondres{\scalebox{0.78}{0.255}} & \scalebox{0.78}{0.221} & \boldres{\scalebox{0.78}{0.254}} & \scalebox{0.78}{0.240} & \scalebox{0.78}{0.271} & \scalebox{0.78}{0.225} & {\scalebox{0.78}{0.259}} & \boldres{\scalebox{0.78}{0.206}} & \scalebox{0.78}{0.277} & \scalebox{0.78}{0.242} & \scalebox{0.78}{0.298} &{\scalebox{0.78}{0.219}} &{\scalebox{0.78}{0.261}}  & \scalebox{0.78}{0.237} &\scalebox{0.78}{0.296} & \scalebox{0.78}{0.261} & \scalebox{0.78}{0.340} & \scalebox{0.78}{0.276} &\scalebox{0.78}{0.336} & \scalebox{0.78}{0.245} &\scalebox{0.78}{0.285}  \\ 
     
					& \scalebox{0.78}{336}  & \boldres{\scalebox{0.78}{0.270}} &\boldres{\scalebox{0.78}{0.296}} &{\scalebox{0.78}{0.273}} &\boldres{\scalebox{0.78}{0.296}} & {\scalebox{0.78}{0.278}} & \boldres{\scalebox{0.78}{0.296}} & \scalebox{0.78}{0.292} & \scalebox{0.78}{0.307} & \scalebox{0.78}{0.278} & \secondres{\scalebox{0.78}{0.297}} & \secondres{\scalebox{0.78}{0.272}} & \scalebox{0.78}{0.335} & \scalebox{0.78}{0.287} & \scalebox{0.78}{0.335} &{\scalebox{0.78}{0.280}} &{\scalebox{0.78}{0.306}} & \scalebox{0.78}{0.283} &\scalebox{0.78}{0.335} & \scalebox{0.78}{0.309} & \scalebox{0.78}{0.378} & \scalebox{0.78}{0.339} &\scalebox{0.78}{0.380} & \scalebox{0.78}{0.321} &\scalebox{0.78}{0.338} \\ 
     
					& \scalebox{0.78}{720}  & {\scalebox{0.78}{0.352}} &{\scalebox{0.78}{0.351}} &{\scalebox{0.78}{0.353}} &{\scalebox{0.78}{0.349}} & \scalebox{0.78}{0.358} & \boldres{\scalebox{0.78}{0.347}} & \scalebox{0.78}{0.364} & \scalebox{0.78}{0.353} & {\scalebox{0.78}{0.354}} & \secondres{\scalebox{0.78}{0.348}} & \scalebox{0.78}{0.398} & \scalebox{0.78}{0.418} & \secondres{\scalebox{0.78}{0.351}} & \scalebox{0.78}{0.386} &\scalebox{0.78}{0.365} &{\scalebox{0.78}{0.359}} & \boldres{\scalebox{0.78}{0.345}} &{\scalebox{0.78}{0.381}} & \scalebox{0.78}{0.377} & \scalebox{0.78}{0.427} & \scalebox{0.78}{0.403} &\scalebox{0.78}{0.428} & \scalebox{0.78}{0.414} &\scalebox{0.78}{0.410}  \\ 
     
					\cmidrule(lr){2-24}
					& \scalebox{0.78}{Avg}  & \boldres{\scalebox{0.78}{0.250}} &\secondres{\scalebox{0.78}{0.278}} &\secondres{\scalebox{0.78}{0.252}} &\boldres{\scalebox{0.78}{0.277}} & {\scalebox{0.78}{0.258}} & {\scalebox{0.78}{0.279}} & \scalebox{0.78}{0.272} & \scalebox{0.78}{0.291} & {\scalebox{0.78}{0.259}} & {\scalebox{0.78}{0.281}} & \scalebox{0.78}{0.259} & \scalebox{0.78}{0.315} & \scalebox{0.78}{0.271} & \scalebox{0.78}{0.320} &{\scalebox{0.78}{0.259}} &{\scalebox{0.78}{0.287}} &\scalebox{0.78}{0.265} &\scalebox{0.78}{0.317} & \scalebox{0.78}{0.292} & \scalebox{0.78}{0.363} &\scalebox{0.78}{0.309} &\scalebox{0.78}{0.360} &\scalebox{0.78}{0.288} &\scalebox{0.78}{0.314} \\ 
				\midrule

                   \multirow{5}{*}{\rotatebox{90}{\scalebox{0.95}{ETTm1}}}
					 &  \scalebox{0.78}{96} & \boldres{\scalebox{0.78}{0.329}}& \boldres{\scalebox{0.78}{0.367}}& \secondres{\scalebox{0.78}{0.331}} & \secondres{\scalebox{0.78}{0.368}}& \secondres{\scalebox{0.78}{0.334}} & \secondres{\scalebox{0.78}{0.368}} & \scalebox{0.78}{0.355} & \scalebox{0.78}{0.376} & \boldres{\scalebox{0.78}{0.329}} & \boldres{\scalebox{0.78}{0.367}} & \scalebox{0.78}{0.404} & \scalebox{0.78}{0.426} & \scalebox{0.78}{0.364} & \scalebox{0.78}{0.387} &{\scalebox{0.78}{0.338}} &{\scalebox{0.78}{0.375}} &{\scalebox{0.78}{0.345}} &{\scalebox{0.78}{0.372}} & \scalebox{0.78}{0.418} & \scalebox{0.78}{0.438} &\scalebox{0.78}{0.379} &\scalebox{0.78}{0.419} &\scalebox{0.78}{0.386} &\scalebox{0.78}{0.398} \\ 
      
                    & \scalebox{0.78}{192} & \scalebox{0.78}{0.388} & \scalebox{0.78}{0.404}& \scalebox{0.78}{0.378} & \scalebox{0.78}{0.393}& \scalebox{0.78}{0.377} & \scalebox{0.78}{0.391} & \scalebox{0.78}{0.391} & \scalebox{0.78}{0.392} & \boldres{\scalebox{0.78}{0.367}} & \boldres{\scalebox{0.78}{0.385}} & \scalebox{0.78}{0.450} & \scalebox{0.78}{0.451} &\scalebox{0.78}{0.398} & \scalebox{0.78}{0.404} &\secondres{\scalebox{0.78}{0.374}} &\secondres{\scalebox{0.78}{0.387}}  &{\scalebox{0.78}{0.380}} &{\scalebox{0.78}{0.389}} & \scalebox{0.78}{0.439} & \scalebox{0.78}{0.450}  &\scalebox{0.78}{0.426} &\scalebox{0.78}{0.441} &\scalebox{0.78}{0.459} &\scalebox{0.78}{0.444}  \\ 
                    
                    & \scalebox{0.78}{336} & \scalebox{0.78}{0.423} & \scalebox{0.78}{0.429}& \scalebox{0.78}{0.410} & \scalebox{0.78}{0.414}& \scalebox{0.78}{0.426} & \scalebox{0.78}{0.420} & \scalebox{0.78}{0.424} & \scalebox{0.78}{0.415} & \boldres{\scalebox{0.78}{0.399}} & \boldres{\scalebox{0.78}{0.410}} & \scalebox{0.78}{0.532}  &\scalebox{0.78}{0.515} & \scalebox{0.78}{0.428} & \scalebox{0.78}{0.425} &\secondres{\scalebox{0.78}{0.410}} &\secondres{\scalebox{0.78}{0.411}}  &{\scalebox{0.78}{0.413}} &{\scalebox{0.78}{0.413}} & \scalebox{0.78}{0.490} & \scalebox{0.78}{0.485}  &\scalebox{0.78}{0.445} &\scalebox{0.78}{0.459} &\scalebox{0.78}{0.495} &\scalebox{0.78}{0.464} \\ 
                    
                    & \scalebox{0.78}{720}& \scalebox{0.78}{0.490} & \scalebox{0.78}{0.461}& \scalebox{0.78}{0.474} & \scalebox{0.78}{0.451} & \scalebox{0.78}{0.491} & \scalebox{0.78}{0.459} & \scalebox{0.78}{0.487} & \scalebox{0.78}{0.450} & \boldres{\scalebox{0.78}{0.454}} & \boldres{\scalebox{0.78}{0.439}} & \scalebox{0.78}{0.666} & \scalebox{0.78}{0.589} & \scalebox{0.78}{0.487} & \scalebox{0.78}{0.461} &{\scalebox{0.78}{0.478}} &\secondres{\scalebox{0.78}{0.450}} &\secondres{\scalebox{0.78}{0.474}} &{\scalebox{0.78}{0.453}} & \scalebox{0.78}{0.595} & \scalebox{0.78}{0.550}  &\scalebox{0.78}{0.543} &\scalebox{0.78}{0.490} &\scalebox{0.78}{0.585} &\scalebox{0.78}{0.516}  \\ 
                    \cmidrule(lr){2-24}
                    & \scalebox{0.78}{Avg}& \scalebox{0.78}{0.408} & \scalebox{0.78}{0.415} & \scalebox{0.78}{0.398} & \scalebox{0.78}{0.407}  & \scalebox{0.78}{0.407} & \scalebox{0.78}{0.410} & \scalebox{0.78}{0.414} & \scalebox{0.78}{0.407} & \boldres{\scalebox{0.78}{0.387}} & \boldres{\scalebox{0.78}{0.400}} & \scalebox{0.78}{0.513} & \scalebox{0.78}{0.496} & \scalebox{0.78}{0.419} & \scalebox{0.78}{0.419} &\secondres{\scalebox{0.78}{0.400}} &\secondres{\scalebox{0.78}{0.406}}  &{\scalebox{0.78}{0.403}} &{\scalebox{0.78}{0.407}} & \scalebox{0.78}{0.485} & \scalebox{0.78}{0.481}  &\scalebox{0.78}{0.448} &\scalebox{0.78}{0.452} &\scalebox{0.78}{0.481} &\scalebox{0.78}{0.456}  \\ 
                    \midrule

					\multicolumn{2}{c|}{\scalebox{0.78}{{$1^{\text{st}}$ Count}}} & \scalebox{0.78}{\boldres{30}} & \scalebox{0.78}{\boldres{30}} &{\scalebox{0.78}{4}} &{\scalebox{0.78}{3}}  & \scalebox{0.78}{{3}} & \scalebox{0.78}{\secondres{7}} & \scalebox{0.78}{0}& \scalebox{0.78}{0}& \scalebox{0.78}{\secondres{5}} & \scalebox{0.78}{5}& \scalebox{0.78}{2}& \scalebox{0.78}{0}& \scalebox{0.78}{0}& \scalebox{0.78}{0}& \scalebox{0.78}{0}& \scalebox{0.78}{0}& \scalebox{0.78}{1}& \scalebox{0.78}{0}& \scalebox{0.78}{0} & \scalebox{0.78}{{0}} & \scalebox{0.78}{0}& \scalebox{0.78}{0}& \scalebox{0.78}{0}& \scalebox{0.78}{0} \\
					
					\bottomrule
				\end{tabular}
		\end{threeparttable}
	}
\end{table*}

As shown in Table \ref{full_baseline_results}, DC-Mamber outperforms existing baselines on multiple datasets, particularly in tasks involving larger data and more variables. For instance, on the PEMS datasets, DC-Mamber significantly outperforms baseline models that adopt only channel-independent or channel-mixing strategies, demonstrating its superior capability in capturing complex dependencies in time series data.

Specifically, in comparison experiments with S-Mamba, DC-Mamber achieves various degrees of performance improvement (original $\rightarrow$ improved values). In terms of MSE: Significant reductions in PEMS03(22.1\%, 0.136$\rightarrow$ 0.106) and
PEMS04(21.3\%, 0.096$\rightarrow$ 0.083). 
More modest but consistent gains are observed in other datasets:
8.0\% (0.088$\rightarrow$ 0.081) in PEMS07,
5.8\% (0.156$\rightarrow$ 0.147) in PEMS08,
1.2\% (0.171$\rightarrow$ 0.169) in Electricity,
4.5\% (0.244$\rightarrow$ 0.233) in Solar-Energy, 
0.8\% (0.252$\rightarrow$ 0.250) in Weather; 
Regarding MAE: The improvements show notable performance on PEMS03 (11.2\%, 0.241$\rightarrow$ 0.214) and PEMS08(14.0\%, 0.243$\rightarrow$ 0.209). Other gains include:
7.8\% (0.204$\rightarrow$ 0.188) in PEMS04,
5.9\% (0.188$\rightarrow$ 0.177) in PEMS07, 
0.8\% (0.267$\rightarrow$ 0.265) in Electricity, 
5.8\% (0.275$\rightarrow$ 0.259) in Solar-Energy.

The superior performance of DC-Mamber over baseline models can be attributed to the following three key factors:
1) DC-Mamber effectively exploits the multi-level feature dependencies inherent in time series data, enabling the modeling of both global temporal dependencies and local intra-variable temporal relationships. In contrast, S-Mamba, although leveraging global information via the Bi-Mamba module, remains limited by Mamba's inherent inability to capture global temporal dependencies. Transformer-based models, while capable of capturing global dependencies, fail to effectively model local dependencies within individual variables;
2) DC-Mamber carefully accounts for the differences between the linear attention mechanism and the Bi-Mamba and designs tailored channel architectures based on channel-independent and channel-mixing strategies. This allows each of DC-Mamber’s core components to fully utilize its strengths in feature processing. Specifically, temporal tokens are processed by the T-Encoder, which employs attention mechanisms to globally model dependencies across all time steps in the sequence, ensuring comprehensive global temporal modeling. Meanwhile, variable tokens are handled by the V-Encoder, which leverages Bi-Mamba's local modeling capability to capture intra-variable temporal dependencies;
3) To mitigate the potential information interference caused by directly applying heterogeneous features to the prediction task, DC-Mamber incorporates a linear Feature-fusion module. This module performs unified modeling and reconstruction of the features extracted by the T-Encoder and V-Encoder, thereby enhancing the cooperative expressiveness of the fused representations.

\begin{figure*}[htbp]
  \centering
  \includegraphics[width=1.02\textwidth]{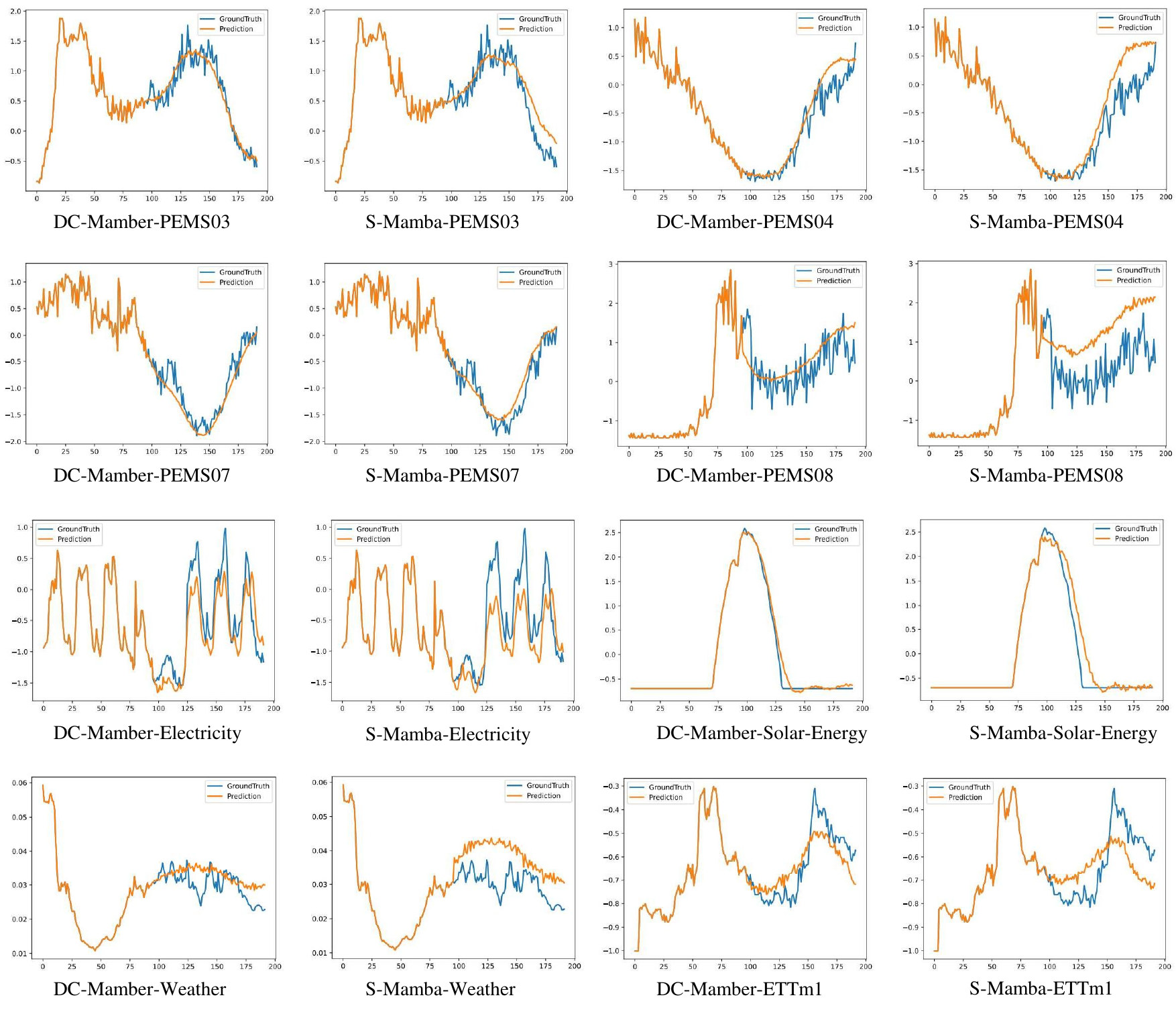} 
  \caption{Comparison of Prediction Performance between DC-Mamber and S-Mamba}
  \label{fig:3-Comparison of forecasts}
\end{figure*}

To visually observe the differences in prediction accuracy between DC-Mamber and baseline models, we visualize the results of DC-Mamber and S-Mamba across eight datasets, as shown in Figure \ref{fig:3-Comparison of forecasts}.


\subsection{Ablation Study}
\begin{figure*}[!htbp]
  \centering
  \includegraphics[width=\textwidth]{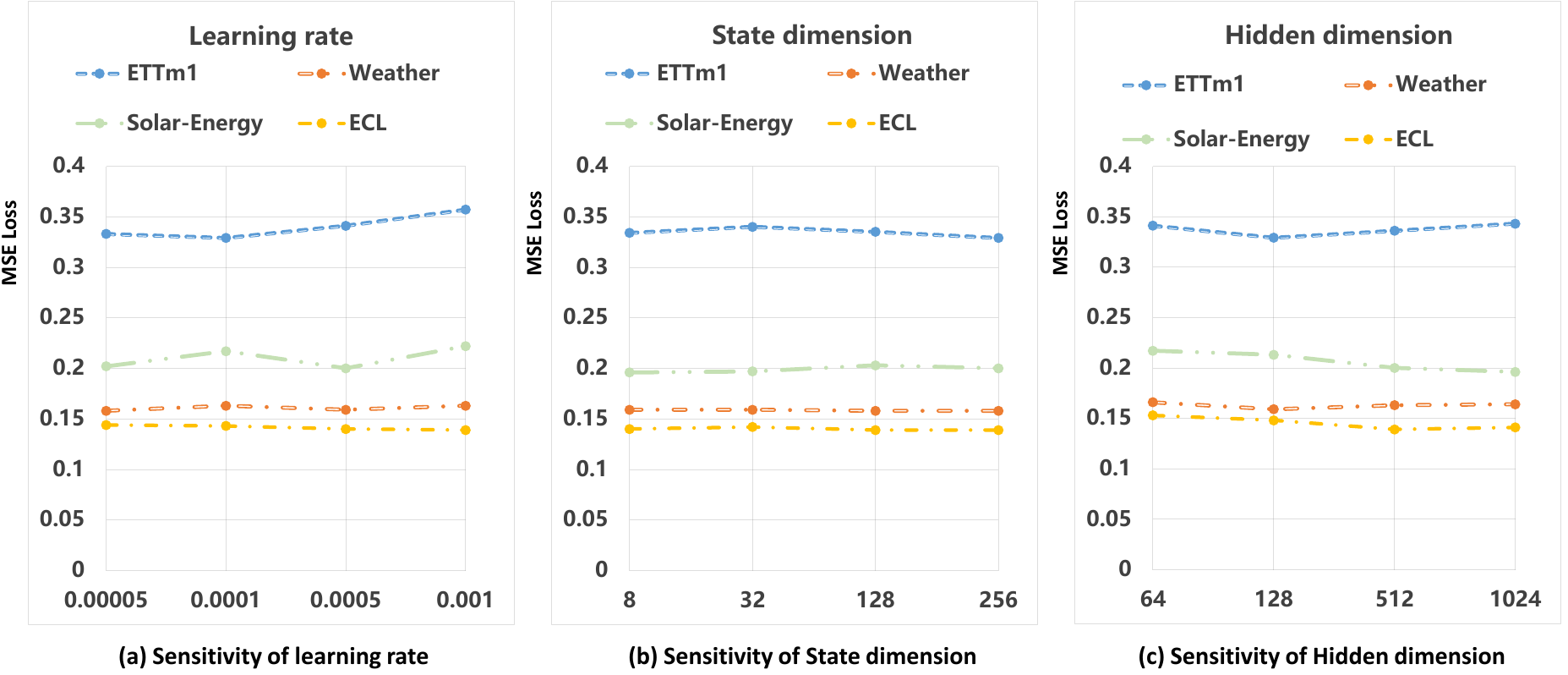} 
  \caption{Sensitivity analysis of hyperparameters on ETTm1, Weather, Solar-Energy, and ECL datasets under a fixed look-back window ($L=96$) and prediction window ($W=96$). }
  \label{fig:2-parameter_sensitivity}
\end{figure*}

To validate the contribution of key components to the model's performance improvements, we conducted five ablation studies designed as follows:
1) Removing individual encoders. We separately removed either the T-Encoder or V-Encoder, forcing the model to perform prediction tasks with only a single-channel encoder, to validate the necessity of the dual-channel encoder architecture.
2) Swapping the roles of encoders. We exchanged the roles of T-Encoder and V-Encoder. In this setting, the T-Encoder—which is originally designed to model temporal dependencies—is assigned to process variable tokens using the channel-independent strategy, while the V-Encoder—which originally captures intra-variable relationships—is tasked with processing temporal tokens using the channel-mixing strategy. This experiment verifies the rationality of aligning the encoder design with the corresponding data processing strategy. In addition, it evaluates the suitability of using the Bi-Mamba module for modeling local intra-variable dependencies and the linear attention mechanism for capturing global temporal dependencies.
3) Using a single data processing strategy for both encoders: While retaining the dual-channel encoder structure, both encoders are fed with data from only one dimension (either temporal or variable), meaning that the model processes data using only the channel-independent or only the channel-mixing strategy. This confirms the benefits of jointly applying both data processing strategies.

The average results of the ablation studies are presented in Table \ref{tab:ablation_comparison_averages} and Table \ref{tab:tokenization_averages}. Detailed ablation results are shown in the Appendix.
From the analysis of the results, several noteworthy conclusions can be drawn:

1) In the single-channel configuration, the predictive performance deteriorates regardless of whether only the channel-mixing T-Encoder or the channel-independent V-Encoder is used. This result demonstrates that the dual-channel structure can capture multi-level temporal features and enhance model performance by separately handling global temporal dependencies and local dependencies among variables.
It confirms the necessity of the dual-channel design.

2) When the roles of the T-Encoder and V-Encoder are swapped, the model exhibits a significant performance drop. This further confirms the validity of designing dedicated channel structures that align with the intrinsic characteristics of the linear attention mechanism and the Bi-Mamba module, in line with the channel-independent and channel-mixing processing strategies. Specifically, It is reasonable to apply the channel mixing strategy to the linear attention-based channels, while employing the channel independence strategy for the Bi-Mamba-based channel architecture design. This design effectively harnesses the respective strengths of attention mechanisms and Mamba in capturing the appropriate types of dependencies.

3) When only a single data processing strategy is employed, the model performance also deteriorates considerably. Especially when solely using the channel-mixing strategy, which results in the poorest performance. This finding highlights that both global temporal dependencies and local dependencies among variables carry essential predictive information, and ignoring either leads to performance degradation. Hence, the results validate the effectiveness of applying both channel-independent and channel-mixing processing strategies.

\begin{table}[ht]
\caption{Ablation average results: Comparing Dual-Encoder and Single-Encoder Architectures in DC-Mamber.}
\label{tab:ablation_comparison_averages}
\centering
\resizebox{\columnwidth}{!}{
\begin{tabular}{c|c|c|cc|cc|cc}
\toprule
\multirow{2}{*}{\textbf{Models}} & \multirow{2}{*}{\textbf{V-Encoder}} & \multirow{2}{*}{\textbf{T-Encoder}} & \multicolumn{2}{c|}{\textbf{PEMS08}} & \multicolumn{2}{c|}{\textbf{Electricity}} & \multicolumn{2}{c}{\textbf{Solar-Energy}} \\
 & & & MSE & MAE & MSE & MAE & MSE & MAE \\ \midrule
\textbf{Ours} & V-Encoder & T-Encoder & \boldres{0.147} & \boldres{0.209} & \boldres{0.169} & \boldres{0.265} & \boldres{0.232} & \boldres{0.258} \\ \midrule
\multirow{2}{*}{w/o} & w/o & T-Encoder & 0.160 & 0.232 & 0.175 & 0.270 & 0.241 & 0.261 \\ \cmidrule(lr){2-9} 
\multirow{2}{*}{} & V-Encoder & w/o & 0.218 & 0.273 & 0.194 & 0.297 & 0.264 & 0.281 \\ \bottomrule
\end{tabular}   }
\end{table}

\begin{table}[ht]
\caption{Ablation average results: Comparison of Tokenization Methods in DC-Mamber.}
\label{tab:tokenization_averages}
\centering
\resizebox{\columnwidth}{!}{
\begin{tabular}{c|c|c|cc|cc|cc} 
\toprule
\multirow{2}{*}{\textbf{Models}} & \multirow{2}{*}{\textbf{V-Encoder}} & \multirow{2}{*}{\textbf{T-Encoder}} & \multicolumn{2}{c|}{\textbf{PEMS04}} & \multicolumn{2}{c|}{\textbf{PEMS08}} & \multicolumn{2}{c}{\textbf{Electricity}} \\
 & & & MSE & MAE & MSE & MAE & MSE & MAE \\ \midrule
\textbf{Ours} & Channel-Independent & Channel-Mixing & \boldres{0.083} & \boldres{0.186} & \boldres{0.147} & \boldres{0.209} & \boldres{0.169} & \boldres{0.265} \\ \midrule
\multirow{4}{*}{Tokenization} & Channel-Mixing & Channel-Independent & 0.089 & 0.197 & 0.151 & 0.217 & 0.225 & 0.318 \\ \cmidrule(lr){2-9} 
\multirow{4}{*}{} & Channel-Independent & Channel-Independent & 0.093 & 0.199 & 0.158 & 0.231 & 0.172 & 0.268 \\ \cmidrule(lr){2-9} 
\multirow{4}{*}{} & Channel-Mixing & Channel-Mixing & 0.152 & 0.260 & 0.203 & 0.259 & 0.188 & 0.290 \\ \bottomrule
\end{tabular}  }
\end{table}

\subsection{Parameter Sensitivity Analysis}

The performance of prediction models is highly influenced by hyperparameter settings, with different models exhibiting varying sensitivity. The specific configurations of DC-Mamber are provided in Table \ref{parameter-pppp} and Table \ref{parameter-eswe}.

We conduct a sensitivity analysis of the model's key hyperparameters to evaluate how different parameter settings affect performance. Specifically, we systematically examine three hyperparameters—Learning Rate ($lr$), State Dimension ($d_{\rm state}$), and Hidden Dimension ($d_{\rm model}$)—across multiple datasets (ETTm1, Weather, Solar-Energy, and ECL).
Experimental results indicate that:  
1) Neither smaller nor larger learning rate ($lr$) values consistently lead to better performance, as different datasets exhibit varying convergence preferences;  
2) The state dimension ($d_{\rm state}$) has a relatively minor impact on prediction accuracy;  
3) Increasing the hidden dimension ($d_{\rm model}$) generally improves performance within a reasonable range, but overly large or small values may degrade it.  

Overall, the model shows low sensitivity to hyperparameter settings—particularly on the Weather and ECL datasets—highlighting the robustness of DC-Mamber.

\section{Conclusion}
This paper proposes a dual-channel time series forecasting model, DC-Mamber, based on linear Transformer and Mamba. Considering the complementary strengths of Transformer and Mamba in capturing global and local features, the corresponding channel structures are designed according to the data processing strategies of channel mixing and channel independence, so that the model can capture both the global temporal dependencies and the local dependencies of variables simultaneously. The dual-channel design addresses the limitations of each individual architecture: the linear attention mechanism compensates for Bi-Mamba’s inability to capture global dependencies, while Bi-Mamba compensates for the linear attention’s weakness in modeling local dependencies. A linear Feature-fusion module helps the model further process the decoupled features extracted by dual- channel, effectively avoiding information confusion. In addition, the computational complexity of each single channel is linear. Our model has been extensively verified on real-world datasets and compared with 11 baselines, which demonstrate the effectiveness of DC-Mamber. The results show that DC-Mamber has achieved SOTA performance.

\bibliographystyle{IEEEtran}
\bibliography{IEEEabrv,ustc2}

\subsection{Appendix}
\begin{table*}[ht]
\caption{The hyperparameters of DC-Mamber on PEMS03, PEMS04, PEMS07, and PEMS08 datasets.}
\label{parameter-pppp}
\renewcommand{\arraystretch}{1.2} 
\centering
\resizebox{\textwidth}{!}{
\begin{threeparttable}
\setlength{\tabcolsep}{4pt}
\begin{tabular}{c|c|cccc|cccc|cccc|cccc}
\toprule
\multicolumn{2}{c|}{Models} & \multicolumn{4}{c|}{PEMS03} & \multicolumn{4}{c|}{PEMS04} & \multicolumn{4}{c|}{PEMS07} & \multicolumn{4}{c}{PEMS08} \\

\cmidrule(lr){1-2}\cmidrule(lr){3-6}\cmidrule(lr){7-10}\cmidrule(lr){11-14}\cmidrule(lr){15-18}
\multicolumn{2}{c|}{Horizon}& 12 & 24 & 48 & 96 & 12 & 24 & 48 & 96 & 12 & 24 & 48 & 96 & 12 & 24 & 48 & 96 \\
\midrule
\multirow{6}{*}{\rotatebox{90}{Hyperparameters}} 
& $el$ & 4 & 4 & 4 & 4      & 4 & 4 & 4 & 4      & 2 & 2 & 4  & 4         & 2 & 2 & 4 & 4 \\

& $bs$ & 32 & 32 & 32 & 32      & 32 & 32 & 32 & 32    & 32 & 32 & 16 & 16  & 32 & 32 & 16 & 16 \\

& $lr$ &5e-4 & 5e-4 & 5e-4 & 5e-4        & 5e-4 & 5e-4 & 5e-4 & 5e-4      & 1e-3 & 1e-3 & 1e-3 & 1e-3      & 5e-4 & 5e-4 & 1e-4 & 1e-4  \\

& $d\_\rm model$ & 512 & 512 & 512 & 512      & 1024 & 1024 & 1024 & 1024     & 512 & 512 & 512 & 512       & 512 & 512 & 512 & 512 \\

& dropout & 0.1 & 0.1 & 0.1 & 0.1      & 0.1 & 0.1 & 0.1 & 0.1          & 0.1 & 0.1 & 0.1 & 0.1        & 0.1 & 0.1 & 0.1 & 0.1 \\

& $d\_\rm state$   & 128 & 128 & 256 & 256   & 256 & 256 & 32 & 32             & 256 & 256 & 256 & 32 & 256 & 256 & 256 & 256 \\
\bottomrule
\end{tabular}
\end{threeparttable}
}
\end{table*}

\begin{table*}[ht]
\caption{The hyperparameters of DC-Mamber on ECL, Solar-Energy, Weather, and ETTm1 datasets.}
\label{parameter-eswe}
\renewcommand{\arraystretch}{1.2} 
\centering
\resizebox{\textwidth}{!}{
\begin{threeparttable}
\setlength{\tabcolsep}{4pt}
\begin{tabular}{c|c|cccc|cccc|cccc|cccc}
\toprule
\multicolumn{2}{c|}{Models} & \multicolumn{4}{c|}{ECL} & \multicolumn{4}{c|}{Solar-Energy} & \multicolumn{4}{c|}{Weather} & \multicolumn{4}{c}{ETTm1} \\

\cmidrule(lr){1-2}\cmidrule(lr){3-6}\cmidrule(lr){7-10}\cmidrule(lr){11-14}\cmidrule(lr){15-18}
\multicolumn{2}{c|}{Horizon}& 96 & 192& 336 & 720 & 96 & 192& 336 & 720 & 96 & 192& 336 & 720& 96 & 192& 336 & 720\\
\midrule
\multirow{6}{*}{\rotatebox{90}{Hyperparameters}} 
& $el$ &3 & 3 & 3 & 3      & 2 & 2 & 2 & 2      & 3 & 3 & 3  & 3        & 2 & 2 & 2 & 2 \\

& $bs$ & 16 & 16 & 16 & 16      & 16 & 16 & 16 & 16     & 32 & 32 & 32 & 32          & 32 & 32 & 32 & 32 \\

& $lr$ & 1e-3 & 1e-3 & 1e-3 & 1e-3         & 5e-4 & 5e-4 & 5e-4 & 5e-4               & 5e-5  & 1e-4 & 1e-3 & 1e-4      & 1e-4 & 1e-4 & 1e-4 & 1e-4  \\

& $d\_\rm model$ & 512 & 512 & 512 & 512      & 512 & 512 & 512 & 512                    & 128 & 512 & 512 & 512       & 128 & 128 & 128 & 128 \\

& dropout & 0.1 & 0.1 & 0.1 & 0.1      & 0.1 & 0.1 & 0.1 & 0.1          & 0.1 & 0.1 & 0.1 & 0.1        & 0.1 & 0.1 & 0.1 & 0.1 \\

& $d\_\rm state$   & 256 & 128 & 256 & 128   & 256 & 256 & 256 & 256                       & 128 & 8 & 128 & 32 & 256 & 256 & 256 & 256 \\
\bottomrule
\end{tabular}
\end{threeparttable}
}
\end{table*}

\begin{table*}[ht]
\caption{Ablation full results: Comparing Dual-Encoder and Single-Encoder Architectures in DC-Mamber.}
\label{ablation_comparison_table}
\renewcommand{\arraystretch}{1.1} 
\centering
\resizebox{\textwidth}{!}{
\begin{threeparttable}
\setlength{\tabcolsep}{3.5pt}
\begin{tabular}{c|c|c|c|cc|c|cc|cc}
\toprule
\multirow{2}{*}{Models} & \multirow{2}{*}{V-Encoder} & \multirow{2}{*}{T-Encoder} & \multirow{2}{*}{Forecasting Lengths} & \multicolumn{2}{c|}{PEMS08} & \multirow{2}{*}{Forecasting Lengths} & \multicolumn{2}{c|}{Electricity} & \multicolumn{2}{c}{Solar-Energy} \\
 & & & & MSE & MAE & & MSE & MAE & MSE & MAE \\
\midrule
\multirow{5}{*}{\centering Ours} 
& \multirow{5}{*}{V-Encoder} 
& \multirow{5}{*}{T-Encoder}
& 12 & \boldres{0.076} & \boldres{0.172} & 96
& \boldres{0.139} & \boldres{0.235} 
& \boldres{0.200} & \boldres{0.228} \\
& & & 24 & \boldres{0.100} & \boldres{0.198} & 192
& \boldres{0.163} & \boldres{0.259} 
& \boldres{0.233} & \boldres{0.256} \\
& & & 48 & \boldres{0.196} & \boldres{0.223} & 336
& \boldres{0.176} & \boldres{0.273} 
& \boldres{0.246} & \boldres{0.273} \\
& & & 96 & \boldres{0.217} & \boldres{0.244} & 720 
& \boldres{0.197} & {0.294} 
& \boldres{0.248} & \boldres{0.274} \\
\cmidrule(lr){4-11}
& & & Avg & \boldres{0.147} & \boldres{0.209} & Avg 
& \boldres{0.169} & \boldres{0.265} 
& \boldres{0.232} & \boldres{0.258} \\
 
\midrule
\multirow{10}{*}{w/o} 
& \multirow{5}{*}{V-Encoder} 
& \multirow{5}{*}{w/o}
& 12 & 0.079 & 0.182 & 96
& 0.140 & 0.237 
& 0.212 & 0.232 \\
& & & 24 & 0.113 & 0.217 & 192 
& 0.172 & 0.265 
& 0.247 & 0.257 \\
& & & 48 & 0.204 & 0.248 & 336 
& 0.192 & 0.285 
& 0.254 & 0.276 \\
& & & 96 & 0.245 & 0.281 & 720 
& \boldres{0.197} & \boldres{0.293} 
& 0.252 & 0.277 \\
\cmidrule(lr){4-11}
& & & Avg & 0.160 & 0.232 & Avg 
& 0.175 & 0.270 
& 0.241 & 0.261 \\
 
\cmidrule(lr){2-11}   
\multirow{5}{*}{} 
& \multirow{5}{*}{w/o} 
& \multirow{5}{*}{T-Encoder}
& 12 & 0.117 & 0.218 & 96 
& 0.176 & 0.281 
& 0.221 & 0.245 \\
& & & 24 & 0.143 & 0.243 & 192 
& 0.187 & 0.290 
& 0.267 & 0.275 \\
& & & 48 & 0.296 & 0.309 & 336
& 0.197 & 0.301 
& 0.265 & 0.292 \\
& & & 96 & 0.316 & 0.322 & 720 
& 0.214 & 0.314 
& 0.303 & 0.312 \\
\cmidrule(lr){4-11}
& & & Avg & 0.218 & 0.273 & Avg 
& 0.194 & 0.297 
& 0.264 & 0.281\\
\bottomrule
\end{tabular}

\end{threeparttable}
}
\end{table*}

\begin{table*}[!ht]
\caption{Ablation full results: Comparison of Tokenization Methods in DC-Mamber.}
\label{ablation_comparison_table2}
\renewcommand{\arraystretch}{1.1} 
\centering
\resizebox{\textwidth}{!}{%
\begin{threeparttable}
\setlength{\tabcolsep}{3.2pt}
\begin{tabular}{c|c|c|c|cc|cc|c|cc}
\toprule
\multirow{2}{*}{Models} & \multirow{2}{*}{V-Encoder} & \multirow{2}{*}{T-Encoder} & \multirow{2}{*}{Forecasting Lengths} & \multicolumn{2}{c|}{PEMS04} & \multicolumn{2}{c|}{PEMS08} & \multirow{2}{*}{Forecasting Lengths} & \multicolumn{2}{c}{Electricity} \\
 & & & & MSE & MAE & MSE & MAE & & MSE & MAE \\
\midrule
\multirow{5}{*}{Ours} 
& \multirow{5}{*}{\begin{tabular}[c]{@{}c@{}}Channel-Independent \end{tabular}} 
& \multirow{5}{*}{\begin{tabular}[c]{@{}c@{}} Channel-Mixing\end{tabular}}
& 12 & \boldres{0.069} & \boldres{0.163} & {0.076} & \boldres{0.172} & 96 & {0.139} & \boldres{0.235} \\
& & & 24 & \boldres{0.076} & \boldres{0.179} & \boldres{0.100} & \boldres{0.198} & 192 & \boldres{0.163} & \boldres{0.259} \\
& & & 48 & \boldres{0.088} & \boldres{0.195} & {0.196} & \boldres{0.223} & 336 & \boldres{0.176} & \boldres{0.273} \\
& & & 96 & \boldres{0.100} & \boldres{0.207} & \boldres{0.217} & \boldres{0.244} & 720 & \boldres{0.197} &{0.294} \\
\cmidrule(lr){4-11}
& & & Avg & \boldres{0.083} & \boldres{0.186} & \boldres{0.147} & \boldres{0.209} & Avg & \boldres{0.169} & \boldres{0.265} \\
 
\midrule
\multirow{15}{*}{Tokenization} 
& \multirow{5}{*}{\begin{tabular}[c]{@{}c@{}} Channel-Mixing\end{tabular}} 
& \multirow{5}{*}{\begin{tabular}[c]{@{}c@{}}Channel-Independent \end{tabular}} 
& 12 & 0.073 & 0.178 & \boldres{0.075 }& 0.176 & 96 & 0.177 & 0.283 \\
& & & 24 & 0.087 & 0.195 & 0.104 & 0.204 & 192 & 0.192 & 0.295 \\
& & & 48 & 0.096 & 0.207 & \boldres{0.195} & 0.229 & 336 & 0.266 & 0.349 \\
& & & 96 & 0.100 & 0.208& 0.231 & 0.258 & 720 & {0.264} & 0.345 \\
\cmidrule(lr){4-11}
& & & Avg & 0.089 & 0.197 & 0.151 & 0.217 & Avg & 0.225 & 0.318 \\
 
\cmidrule(lr){2-11}   
\multirow{5}{*}{} 
& \multirow{5}{*}{\begin{tabular}[c]{@{}c@{}}Channel-Independent \end{tabular}} 
& \multirow{5}{*}{\begin{tabular}[c]{@{}c@{}}Channel-Independent \end{tabular}} 
& 12 & 0.072 & 0.176 & 0.077 & 0.179 & 96 & \boldres{0.138} & \boldres{0.235 }\\
& & & 24 & 0.084 & 0.190 & 0.107 & 0.211 & 192 & 0.165 & \boldres{0.259} \\
& & & 48 & 0.098 & 0.206 & 0.197 & 0.246 & 336 & 0.182 & 0.278 \\
& & & 96 & 0.118 & 0.225 & 0.249 & 0.288 & 720 & 0.202 & 0.299 \\
\cmidrule(lr){4-11}
& & & Avg & 0.093 & 0.199 & 0.158 & 0.231 & Avg & 0.172 & 0.268 \\
 
\cmidrule(lr){2-11}   
\multirow{5}{*}{} 
& \multirow{5}{*}{\begin{tabular}[c]{@{}c@{}}  Channel-Mixing\end{tabular}} 
& \multirow{5}{*}{\begin{tabular}[c]{@{}c@{}} Channel-Mixing \end{tabular}} 
& 12 & 0.260 & 0.368 & 0.112 & 0.213 & 96 & 0.170 & 0.274 \\
& & & 24 & 0.112 & 0.219 & 0.141 & 0.238 & 192 & 0.180 & 0.282 \\
& & & 48 & 0.117 & 0.225 & 0.283 & 0.292 & 336 & 0.190 & 0.292 \\
& & & 96 & 0.118 & 0.226 & 0.277 & 0.293 & 720 & 0.211 & 0.312 \\
\cmidrule(lr){4-11}
& & & Avg & 0.152 & 0.260 & 0.203 & 0.259 & Avg & 0.188 & 0.290 \\
\bottomrule
\end{tabular}

\end{threeparttable}%
}
\end{table*}

\end{document}